\documentclass[preprint,12pt]{elsarticle}

\usepackage{lmodern}
\usepackage{amssymb}
\usepackage{amsmath}
\usepackage{booktabs}
\usepackage{array}
\usepackage{multirow}
\usepackage{graphicx}
\usepackage{orcidlink}
\graphicspath{{figures/}}

\newcommand{\repourl}{https://github.com/Cad-Kernel/MIRAGE-CAD}

\usepackage{subcaption}
\usepackage{xcolor}
\usepackage{listings}
\usepackage{algorithm}
\usepackage{algpseudocode}
\usepackage{chngcntr}
\usepackage{hyperref}
\hypersetup{hidelinks}

\definecolor{codeblue}{rgb}{0.1,0.2,0.8}
\makeatletter
\def\ps@pprintTitle{%
  \let\@oddhead\@empty
  \let\@evenhead\@empty
  \def\@oddfoot{\reset@font\hfil\thepage\hfil}%
  \let\@evenfoot\@oddfoot}
\makeatother

\begin{document}

\begin{frontmatter}

  \title{MIRAGE-CAD: Construction-Mediated Multimodal Generation of
    Executable CAD Programs}

  \author{Jizong Zhan~\orcidlink{0009-0006-3762-0328}}
  \ead{jizongzh@gmail.com}

  \begin{abstract}
    Recovering an executable parametric CAD program from an observed object is
    fundamentally ambiguous, because the same final geometry can result from
    different construction procedures. We study this problem from four types of
    input: natural-language descriptions, rendered images, point clouds, and
    STEP/B-Rep geometry. MIRAGE-CAD maps each input to a shared construction representation and
    mediates program generation through an explicit construction-plan interface.
    The resulting Python CAD code is executed by an OpenCASCADE kernel to build
    the solid and export it as STEP. On $2{,}500$ held-out queries per
    modality, the system achieves $55.4$--$70.0\%$ build success and
    $52.3$--$66.2\%$ STEP export success without retrieval at inference. Controlled
    comparisons show that strong reconstruction does not depend on expressing the
    construction representation as text: a decoder conditioned directly on the
    continuous representation also reconstructs strongly, while an
    exposure-matched plan-based decoder shows no detected material loss in per-part
    geometric fidelity. The explicit plan instead provides a readable and
    separately measurable intermediate representation whose agreement with the
    reference construction is informative about downstream execution success.
    Finally, we show that executable validity, geometric fidelity, and parametric
    responsiveness can diverge substantially and should therefore be evaluated
    separately.
  \end{abstract}


\end{frontmatter}

\section{Introduction}
\label{sec:intro}

\subsection{Shape is not construction}
\label{sec:shape_not_construction}

A history-based parametric CAD model contains more than its final shape. It
also records a construction procedure composed of sketches, geometric
operations, dependencies, and parameters. This construction determines how the
model responds to later edits and how design variants can be produced. A mesh
or a history-free boundary representation (B-Rep) may preserve the final
geometry accurately, but it does not by itself preserve the feature
dependencies and parameter semantics needed for history-based editing.

The difficulty is that construction is not uniquely determined by shape. A
cylinder, for example, may be created by extruding a circle, revolving a
rectangle, or cutting a cylindrical pocket through a block. These procedures
can produce the same final solid within geometric tolerance while responding
differently when their parameters are changed. Recovering construction from an
observation is therefore underdetermined: geometric agreement with the final
part does not identify a unique feature history.

A final observation consequently cannot determine \emph{the} construction
history that produced it. CAD reverse engineering can instead aim to recover
\emph{a} construction that is consistent with the observation and useful for
subsequent editing. Existing approaches generate such constructions either as
CAD command sequences
\cite{wu2021deepcad,xu2022skexgen,jayaraman2023solidgen}
or as executable code
\cite{rukhovich2025cadrecode,guan2025cadcoder,wang2025cadfusion}.
Conditioning has also expanded beyond point clouds to text
\cite{khan2024text2cad}, images \cite{you2025img2cad}, and multimodal inputs
\cite{kolodiazhnyi2025cadrille}. In most cases, however, the construction
sequence is treated as the final prediction. This leaves open a different
question: what is gained when construction is introduced as an intermediate
representation before executable CAD code is generated?

\subsection{Retrieval is strong but finite}
\label{sec:retrieval_limits}

Retrieval provides a strong alternative when a corpus already contains
structurally similar construction programs. A query can be matched to a
training example and its construction reused by the code generator, following
the general principle of retrieval-augmented generation
\cite{lewis2020rag} and nearest-neighbour modelling
\cite{khandelwal2020knnlm}. On procedural CAD data, this can work particularly
well because different template families may still share the same or very
similar operation skeletons.

Retrieval nevertheless has a different scope from query-conditioned
generation. It can only reuse constructions represented in the corpus, and the
query affects the output primarily through neighbour selection. A retrieved
construction may therefore yield a valid CAD program while still differing
substantially from the observed part in geometric properties such as scale.
We consequently use retrieval as a strong reference rather than as the target
of the proposed method. Our focus is on generating a construction
representation from the query itself and on understanding what that
representation contributes to the final CAD program.

\subsection{Hypothesis and approach}
\label{sec:hypothesis}

We study two related questions. First, does mediating CAD program generation
through a construction representation help compared with mapping the query
directly to code? Second, if such a representation is useful, what additional
value is obtained by expressing it as an explicit textual construction plan?

\begin{figure}[htbp]
  \centering
  \includegraphics[width=\textwidth]{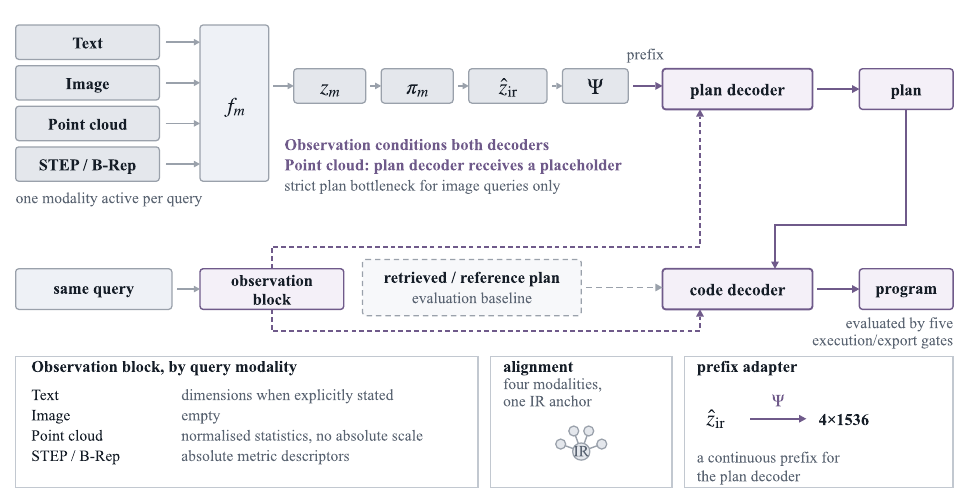}
  \caption{\textbf{Overview of the MIRAGE-CAD architecture.}
    One input modality is active for each query. Its encoder $f_m$ produces
    $z_m$, which is mapped by the modality-specific prior $\pi_m$ to the
    predicted construction latent $\hat{z}_{\mathrm{ir}}$. The prefix adapter
    $\Psi$ converts this latent into a continuous prefix for the plan decoder,
    and the generated plan is then passed to the code decoder to produce an
    executable Python CAD program. A query-derived observation block also
    conditions the decoders. It is empty for image queries, so the plan forms
    a strict information bottleneck in that case. Text and STEP/B-Rep queries
    retain additional observation information, while point-cloud queries use
    a placeholder at the plan stage and normalised statistics at the code
    stage. The dashed retrieved/reference-plan branch is used only for
    evaluation and is not part of deployed generation.}
  \label{fig:architecture}
\end{figure}

To study these questions, we introduce \textbf{MIRAGE-CAD}, a multimodal CAD
program generation pipeline for natural-language descriptions, rendered
images, point clouds, and STEP/B-Rep models. Each modality is encoded
separately and mapped to a shared Construction Intermediate Representation
(IR). A modality-specific prior predicts the corresponding construction
latent, which is transformed into a short continuous prefix for a shared plan
decoder. The decoder produces a textual construction plan, and a second shared
decoder converts that plan into executable Python CAD code. The generated
program is then executed by an OpenCASCADE geometric kernel.

The plan is not the only information available to the code decoder for every
modality. A modality-dependent observation block supplies additional
query-derived information to both the plan and code stages. For image queries
this block is empty, so the generated plan forms a strict information
bottleneck between the image and the program. Text, point-cloud, and STEP
queries retain an auxiliary observation channel. We make this distinction
explicit because it limits what can be attributed to plan mediation alone.

This formulation allows construction conditioning and textualisation to be
studied separately. We compare direct query-to-code generation, direct
conditioning on the continuous construction representation, explicit
textual-plan-mediated generation, retrieval-assisted variants, and
interventions on the information supplied to the decoders. We also separate three properties
that are often conflated in CAD generation. A program may execute successfully
without reconstructing the correct geometry, and a geometrically plausible
program may expose only limited parameter control. We therefore evaluate
executable validity, geometric fidelity, and parametric behaviour as distinct
properties of the generated CAD program.

\subsection{Contributions}
\label{sec:contributions}

The main contributions of this work are:

\begin{itemize}

  \item \textbf{Multimodal CAD generation through a shared construction
          representation.}
        MIRAGE-CAD supports text, image, point-cloud, and STEP/B-Rep queries with
        modality-specific encoders followed by shared plan and code decoders. The
        generated Python programs are executed by an OpenCASCADE kernel and can
        produce solid models and STEP output without retrieval at inference.

  \item \textbf{A controlled study of construction conditioning and
          textualisation.}
        We separate conditioning on a continuous construction representation from
        expressing that representation as text. The experiments show that strong
        reconstruction does not require textualisation, while the explicit plan
        provides an inspectable and separately measurable interface. In the
        exposure-matched comparison, no material loss in per-part geometric fidelity
        is detected when the construction representation is expressed as a plan.

  \item \textbf{CAD-native evaluation of validity, fidelity, and parametric
          behaviour.}
        Executable validity is assessed through sequential parsing, execution,
        solid-building, kernel-validity, and STEP-export gates. It is reported
        separately from geometric fidelity and from parametric behaviour, the latter
        characterised through rebuild validity, geometric response to parameter
        perturbations, and parameter reachability.

  \item \textbf{Analysis of retrieval redundancy and external model
          boundaries.}
        We analyse how cross-family construction-skeleton redundancy contributes to
        the strength of retrieval on procedural CAD data. We further position
        MIRAGE-CAD against the released specialised CAD-Recode model
        \cite{rukhovich2025cadrecode} on externally authored Fusion 360 parts
        \cite{willis2021fusion360gallery}, showing why measurable-geometry coverage
        and geometric fidelity must be considered separately.

\end{itemize}

\section{Related Work}
\label{sec:related}

\subsection{Parametric CAD program generation}

Recent work on parametric CAD generation differs mainly in how construction is
represented.

\textbf{CAD-specific command sequences.}
DeepCAD introduced autoregressive generation of sketch-and-extrude sequences
\cite{wu2021deepcad}, followed by work on disentangled latent spaces
\cite{xu2022skexgen}, hierarchical representations
\cite{xu2023hierarchicalcad}, and direct generation of boundary
representations
\cite{jayaraman2023solidgen,xu2024brepgen,lee2025brepdiff}.
A fixed CAD vocabulary constrains the syntax of the output, but geometric
validity still depends on predicted parameters and operation ordering. Its
expressiveness is also limited to the operations represented by that
vocabulary.

\textbf{Executable CAD code.}
A second line generates Python or another executable modelling language.
CAD-Recode reconstructs CadQuery programs from point clouds
\cite{rukhovich2025cadrecode}. Other systems generate CAD programs from text
using geometric reasoning or chain-of-thought supervision
\cite{guan2025cadcoder}, instruction tuning
\cite{govindarajan2025cadmium,li2025cadllama}, explicit controllability
\cite{zhang2025flexcad,zhang2025geocad}, or reinforcement learning
\cite{yin2026rlcad}. Host-language generation provides access to a richer CAD
API, but it also introduces syntax, name-resolution, and API-level failures.
For this reason, executable-code methods require explicit execution-based
validation in addition to sequence-level evaluation.

\textbf{Conditioning from different input modalities.}
CAD generation has also expanded beyond point clouds to text
\cite{khan2024text2cad}, rendered images
\cite{you2025img2cad,alam2025gencad}, engineering drawings
\cite{wang2025cad2program,tang2025chatcad}, sketches
\cite{wu2024cadvlm}, and multimodal systems
\cite{kolodiazhnyi2025cadrille,zhou2025multimodalcadsurvey}.
For B-Rep input, geometry can be encoded through UV parameterisations
\cite{jayaraman2021uvnet} or topology-aware graph representations
\cite{lambourne2021brepnet,colligan2022hierarchicalcadnet}.
Our STEP branch instead uses compact descriptors extracted from the geometric
kernel.

Across these lines of work, the construction sequence is usually treated as
the final prediction. MIRAGE-CAD instead uses construction as an intermediate
target between the input observation and the executable CAD program.

\subsection{Multimodal conditioning and intermediate representations}

Non-textual observations can reach a language model in several ways.
One option is direct continuous conditioning, in which an encoder output is
projected into the model's embedding space through learned query modules,
linear projectors, or soft prompts
\cite{li2023blip2,liu2023llava,li2021prefixtuning,lester2021prompttuning}.
This preserves dense information, but the conditioning itself is not directly
inspectable.

A second option is retrieval. Similar training examples can be selected and
used as additional context
\cite{lewis2020rag,khandelwal2020knnlm}. Retrieval over multimodal
representations has also been used outside generative modelling to transfer
information from content-similar examples; Chen et al.~\cite{chen2026denoisingimplicitfeedbackcoldstart},
for example, retrieve multimodally similar warm items to construct supervision
for cold-start recommendation. In CAD, retrieval has been used to match and
align existing models to images \cite{gao2024diffcad}. Retrieved examples are
easy to inspect, but their usefulness depends on what relevant structures are
already represented in the corpus.

A third option is to generate an explicit intermediate structure before the
final output. Examples include chains of thought
\cite{wei2022cot}, plans \cite{jiang2024selfplanning}, structured outlines
\cite{li2025scot}, and executable intermediate programs
\cite{gao2023pal}. Such representations can be inspected and evaluated
separately, but they force part of the conditioning through an explicit
interface.

MIRAGE-CAD combines continuous conditioning with an explicit construction
plan. A continuous construction latent is predicted in a shared IR-aligned
space and used to condition the plan decoder; the resulting plan is then
passed to the code decoder. The shared latent space follows the idea of
single-anchor multimodal alignment used in ImageBind
\cite{girdhar2023imagebind}, while the anchor here is construction rather than
an appearance modality. This design makes it possible to compare continuous
construction conditioning with its textualised counterpart directly.

\subsection{Evaluating generated CAD}

Evaluation of generated CAD usually considers sequence agreement, execution,
and geometry, but these measures answer different questions.

\textbf{Sequence similarity} compares generated commands with a reference
construction. It is inexpensive, but alternative command sequences may produce
the same final geometry, so disagreement with one reference does not
necessarily imply geometric error.

\textbf{Execution} tests whether generated code runs and produces a valid CAD
object, following the broader practice of executable evaluation in code
generation
\cite{chen2021codex,li2022alphacode}. In CAD, execution establishes that a
program is syntactically and geometrically admissible to the kernel, but it
does not by itself establish agreement with the target shape.

\textbf{Geometric fidelity} measures that agreement using quantities such as
Chamfer distance, F-score, or volumetric IoU. Code-generating CAD systems
therefore commonly report geometry alongside validity
\cite{rukhovich2025cadrecode}.

\textbf{Parametric behaviour} is less standardised. Prior work studies
editable reconstruction
\cite{zhang2025ecadnet,kodnongbua2023reparamcad} and history-aware modelling
\cite{dong2026histcad}, but fewer evaluations test whether exposed parameters
actually control the generated geometry. We therefore measure rebuild
validity after perturbation, whether the geometry changes, and how much of the
program's geometric numeric content is reachable through declared parameters.

Procedural benchmarks introduce an additional concern: similar construction
patterns may recur across nominally different data partitions. Related work on
data duplication and nearest-neighbour reuse has shown how such redundancy can
affect evaluation
\cite{lee2022dedup,khandelwal2020knnlm}. We therefore measure construction
redundancy in our own corpus and complement the in-distribution experiments
with externally authored Fusion 360 parts
\cite{willis2021fusion360gallery}.

\section{Problem Formulation}
\label{sec:formulation}

\subsection{Inputs and output}
\label{sec:formulation_io}

We consider the generation of a single-part parametric CAD program from one
input modality. Let
\begin{equation}
  m \in \mathcal{M}
  =
  \{\text{text},\ \text{image},\ \text{point},\ \text{step}\},
\end{equation}
and let $x_m$ denote the corresponding query. The four modalities are a
natural-language description, a rendered image, a surface point cloud
containing 3D positions, and a descriptor extracted from STEP/B-Rep geometry.
Exactly one modality is provided for each query; multimodal fusion is not
considered in this study.

The output is an executable Python CAD program rather than a geometric object
directly. Let $\mathcal{Y}$ be the set of candidate program strings and
$\mathcal{B}$ the set of B-Rep solids. Program execution is represented by the
partial map
\begin{equation}
  \label{eq:kernel}
  \mathcal{K}_{\mathrm{exec}} :
  \mathcal{Y} \rightharpoonup \mathcal{B}.
\end{equation}
The map is partial because a generated program may fail to parse, fail during
execution, or execute without producing a valid solid.

For geometric evaluation, a successfully produced solid is converted into an
evaluator-specific representation,
\begin{equation}
  \mathcal{E}_{e} :
  \mathcal{B} \rightharpoonup \mathcal{R}_{e},
  \qquad
  e \in \{\mathrm{int},\mathrm{ext}\}.
\end{equation}
The internal evaluator samples the B-Rep surface directly, whereas the external
evaluator re-tessellates the exported STEP model. The internal sampling and
scoreability protocol is specified in Section~\ref{sec:eval_protocols}, while
the external evaluation protocol is described in
Section~\ref{sec:external_protocol}.

\subsection{Construction intermediate representation}
\label{sec:formulation_ir}

MIRAGE-CAD introduces a Construction Intermediate Representation (IR) between
the query and the final program. An IR document $\mathbf{r}$ contains a header,
a set of parameter declarations, an ordered sequence of feature records, and a
terminator:
\begin{equation}
  \mathbf{r}
  =
  \left\langle
  h,\,
  (p_1,\ldots,p_{n_p}),\,
  (F_1,\ldots,F_{n_f}),\,
  \texttt{END}
  \right\rangle .
\end{equation}

Each feature record identifies a CAD operation and may reference named
parameters and dependencies. The operation vocabulary contains 44
\texttt{OP\_}-prefixed tokens. Parameters may be referenced symbolically, for
example through \texttt{@hole\_radius}, rather than copied into features as
numeric literals. This distinction later allows us to test whether exposed
parameters actually control the generated geometry. The complete grammar and
operation inventory are given in Appendix~\ref{app:ir_grammar}.

The IR records construction structure rather than program implementation
details. Imports, helper functions, variable names, and general control flow
remain part of the generated Python program, not the IR.

\subsection{Task factorisation}
\label{sec:task_factorisation}

For a query $x_m$, generation is factorised as
\begin{equation}
  x_m
  \xrightarrow{f_m}
  z_m
  \xrightarrow{\pi_m}
  \hat{z}_{\mathrm{ir}}
  \xrightarrow{\Psi,\,G_{\mathrm{ir}}}
  \hat{r}
  \xrightarrow{G_{\mathrm{code}}}
  \hat{y}
  \xrightarrow{\mathcal{K}_{\mathrm{exec}}}
  \hat{B}.
\end{equation}

The modality encoder $f_m$ produces an embedding $z_m$. A modality-specific
prior $\pi_m$ maps this embedding to a predicted construction latent
$\hat{z}_{\mathrm{ir}}$. The prefix adapter $\Psi$ converts that latent into a
continuous prefix used by the plan decoder $G_{\mathrm{ir}}$ to generate the
construction plan $\hat{r}$. The code decoder $G_{\mathrm{code}}$ then produces
the executable program $\hat{y}$.

Both decoders may also receive a modality-dependent observation block
$c_{\mathrm{obs}}$ derived from the same query. This block is empty for image
queries. For image input, the plan is therefore a strict information
bottleneck between the query and the code decoder; the other modalities retain
an auxiliary observation channel.

\subsection{Evaluation axes}
\label{sec:evaluation_axes}

We evaluate a generated program along three separate axes.

\textbf{Executable validity.}
Let $V(y) \in \{0,1\}$ denote five-gate validity. A program is valid only if it
(1) parses as Python, (2) executes successfully, (3) builds a solid,
(4) passes kernel validation, and (5) successfully exports a STEP file. These
gates describe whether the program is executable and exportable; they do not
measure whether the reconstructed geometry is correct.

\textbf{Geometric fidelity.}
For outputs whose geometry can be extracted by the relevant evaluator, fidelity
is measured against the target geometry using Chamfer distance and F-score;
volumetric IoU is additionally used in the external evaluation. The exact
sampling procedure, thresholds, and self-sampling calibration are specified in
Section~\ref{sec:eval_protocols}.

\textbf{Parametric behaviour.}
Let $\theta$ denote the declared numeric parameters of a generated program and
let $y[\theta']$ be the same program after changing a tested parameter value.
We measure three quantities: whether the modified program remains five-gate
valid, whether the resulting geometry changes, and how much of the program's
geometric numeric content is reachable through declared parameters. The first
two are measured over a finite set of parameter perturbations; reachability
describes the scope over which those perturbation results apply.

Validity, fidelity, and parametric behaviour are reported separately because
they describe different properties of the generated program.

\subsection{Scope and assumptions}
\label{sec:formulation_scope}

The study is restricted to single connected solids. Assemblies, mating
constraints, and full 2D sketch-constraint graphs are outside its scope.
Training supervision comes from procedurally generated, kernel-validated
programs~\cite{zhan2026fllumaonecodenativemultimodalcad}. Finally, a final
geometry does not uniquely identify its original construction history.
Accordingly, the objective is to recover a plausible construction consistent
with the observation, rather than to claim recovery of the unique original
modelling history.

\section{MIRAGE-CAD}
\label{sec:method}

\subsection{Architecture overview}
\label{sec:arch_overview}

MIRAGE-CAD maps a query from one of four input modalities to an executable
Python CAD program. The generation path is

\begin{equation}
  x_m
  \xrightarrow{f_m}
  z_m
  \xrightarrow{\pi_m}
  \hat{z}_{\mathrm{ir}}
  \xrightarrow{\Psi}
  C_m
  \xrightarrow{G_{\mathrm{ir}}}
  \hat{r}
  \xrightarrow{G_{\mathrm{code}}}
  \hat{y}
  \xrightarrow{\mathcal{K}_{\mathrm{exec}}}
  \hat{b}
  \xrightarrow{\mathrm{export}}
  \mathrm{STEP}.
\end{equation}

Figure~\ref{fig:architecture} summarises the architecture. The modality
encoder $f_m$ produces an embedding $z_m$, and a modality-specific prior
$\pi_m$ maps it to a predicted Construction-IR latent
$\hat{z}_{\mathrm{ir}}$. The prefix adapter $\Psi$ converts this latent into
continuous vectors $C_m$ for the plan decoder $G_{\mathrm{ir}}$, which
generates a textual construction plan $\hat{r}$. A second decoder
$G_{\mathrm{code}}$ generates the Python CAD program $\hat{y}$, which is
executed by the Flluma/OpenCASCADE kernel to obtain a B-Rep solid and, if all
validity gates are passed, a STEP file.

Only one modality encoder and one modality-specific prior are active for a
given query. The prefix adapter, plan decoder, code decoder, and kernel path are
shared across modalities. The deployed generation path does not consult a
retrieval index.

The construction latent and the query observation follow different routes.
The latent $\hat{z}_{\mathrm{ir}}$ conditions the plan decoder only; it reaches
the code decoder through the generated plan. In parallel, a
modality-dependent observation block $c_{\mathrm{obs}}$ is supplied to the
decoders. This block is empty for image queries, making the plan a strict
information bottleneck in that case. Text queries retain the query
description, STEP queries retain kernel-derived geometric statistics, and
point-cloud queries provide normalised statistics to the code stage. For the
point-cloud plan stage, a constant placeholder is used instead. This
asymmetry is kept unchanged in all experiments reported in this paper.

\begin{table}[htbp]
  \centering
  \small
  \caption{Query-derived observation information used by the decoders.}
  \label{tab:observation}
  \begin{tabular}{
      l
      >{\raggedright\arraybackslash}p{0.50\linewidth}
      >{\raggedright\arraybackslash}p{0.22\linewidth}
    }
    \toprule
    Modality & Observation content                      & Absolute scale \\
    \midrule
    Text
             & Query description (up to 400 characters)
             & When explicitly stated                                    \\
    Image
             & Empty
             & No                                                        \\
    Point cloud
             & Normalised shape and PCA statistics
             & No                                                        \\
    STEP/B-Rep
             & Kernel-derived geometric statistics
             & Yes                                                       \\
    \bottomrule
  \end{tabular}
\end{table}

\subsection{Multimodal construction representation}
\label{sec:multimodal_rep}

\subsubsection{Modality encoders}
\label{sec:encoders}

Each query modality has its own encoder, followed by a projection head that
maps its output to a common 512-dimensional unit-normalised embedding space.

\textbf{Text.}
Natural-language descriptions are encoded with a frozen
DistilBERT~\cite{sanh2020distilbert}. Because DistilBERT has no dedicated
pooler, token embeddings are combined by attention-masked mean pooling before
projection.

\textbf{Image.}
Rendered images are encoded with the frozen CLIP ViT-B/32 vision
tower~\cite{radford2021clip}. The experiments use one $224\times224$
isometric render per part.

\textbf{Point cloud.}
Point clouds are encoded by a lightweight
PointNet~\cite{qi2017pointnet} operating on $1{,}024$ surface points. The
input contains xyz coordinates only. Each cloud is centred and divided by its
maximum radius before encoding, so this branch does not retain absolute metric
scale.

\textbf{STEP/B-Rep.}
The STEP branch uses descriptors computed by the geometric kernel rather than
a learned topology encoder. This differs from B-Rep representations based on
UV parameterisation~\cite{jayaraman2021uvnet} or graph message
passing~\cite{lambourne2021brepnet}. The implemented encoder supports global,
face, edge, and relation streams, but the corpus used in this work contains no
independent per-face or per-edge descriptor records. Consequently, the
reported STEP results are effectively produced from a 50-dimensional global
descriptor containing topology and metric statistics. The complete descriptor
layout is given in Appendix~\ref{app:step_features}.

All query encoders use the same projection-head structure and produce
$L^2$-normalised embeddings
\[
  z_m \in \mathbb{S}^{511},
\]
so dot products correspond to cosine similarity.

\subsubsection{IR-anchored alignment}
\label{sec:alignment}

The four query modalities are aligned to the Construction IR rather than to
one another. The reference IR $\mathbf{r}$ is encoded by an additional
DistilBERT-based encoder $f_{\mathrm{ir}}$ using the same projection head,
giving $z_{\mathrm{ir}} \in \mathbb{S}^{511}$. Alignment is trained with

\begin{equation}
  \mathcal{L}_{\mathrm{align}}
  =
  \frac{1}{4}
  \sum_{m\in\mathcal{M}}
  \mathcal{L}_{\mathrm{InfoNCE}}
  (z_m,z_{\mathrm{ir}}),
\end{equation}

using symmetric InfoNCE with temperature $\tau=0.07$.

This produces a star-shaped alignment space with the Construction IR as the
common anchor. The same general single-anchor idea appears in
ImageBind~\cite{girdhar2023imagebind}, although the anchor here represents
construction rather than visual appearance. We use this topology as a design
choice; the paper does not compare it experimentally with full pairwise
alignment.

\subsubsection{Construction prior}
\label{sec:priors}

At inference, the reference construction $\mathbf{r}$ and therefore
$z_{\mathrm{ir}}$ are unavailable. A separate prior for each modality predicts
the corresponding position in construction space:

\begin{equation}
  \hat{z}_{\mathrm{ir}}
  =
  \pi_m(z_m),
  \qquad
  \pi_m :
  \mathbb{S}^{511}
  \rightarrow
  \mathbb{S}^{511}.
\end{equation}

Each $\pi_m$ is a residual MLP followed by $L^2$ normalisation. The four priors
are trained independently while the alignment encoders remain frozen. Their
objective combines Euclidean, cosine, and contrastive terms:

\begin{equation}
  \mathcal{L}_{\mathrm{prior}}
  =
  \lambda_{L2}\mathcal{L}_{L2}
  +
  \lambda_{\mathrm{cos}}\mathcal{L}_{\mathrm{cos}}
  +
  \lambda_{\mathrm{nce}}\mathcal{L}_{\mathrm{InfoNCE}},
\end{equation}

with all three weights set to $1.0$ in the reported experiments.

\subsection{Construction-plan mediation}
\label{sec:plan_mediation}

\subsubsection{Prefix adapter}
\label{sec:soft_prefix}

The predicted construction latent is continuous and has no token identity.
Rather than mapping it to the nearest textual IR, MIRAGE-CAD uses it directly
as conditioning. The prefix adapter produces

\begin{equation}
  C_m
  =
  \Psi(\hat{z}_{\mathrm{ir}})
  \in
  \mathbb{R}^{K\times H},
  \qquad
  K=4,\quad H=1536.
\end{equation}

The adapter consists of LayerNorm, a linear projection to the decoder
dimension, GELU activation, and a second linear layer producing the
$K\times H$ output. The resulting vectors have no vocabulary tokens. They are
a query-dependent continuous prefix, related to prefix
tuning~\cite{li2021prefixtuning} but predicted from the input rather than
learned as fixed free parameters.

In the deployed architecture, this prefix conditions only the plan decoder.
The code decoder receives no copy of $\hat{z}_{\mathrm{ir}}$ or $C_m$.

\subsubsection{Plan decoder}
\label{sec:plan_decoder}

The plan decoder $G_{\mathrm{ir}}$ is based on
Qwen2.5-Coder-1.5B~\cite{yang2025qwen25}. The base model is loaded in 4-bit
quantised form following QLoRA~\cite{dettmers2023qlora} and adapted with
LoRA~\cite{hu2021lora}. The prefix vectors are inserted directly before the
token embeddings,

\begin{equation}
  e_{\mathrm{in}}
  =
  \left[
    C_m;
    E_{\mathrm{tok}}(\mathrm{prompt})
    \right]
  \in
  \mathbb{R}^{(K+T)\times H}.
\end{equation}

The prompt contains a short task instruction, the query modality, the
available observation block $c_{\mathrm{obs}}$, and an output cue. It does not
contain the Construction-IR grammar. During training, the target reference IR
$\mathbf{r}$ follows the prompt, but the loss is applied only to target plan
tokens; prefix and prompt positions are masked. At inference, the decoder
receives the prefix and prompt and generates $\hat{r}$ autoregressively.

\subsection{Plan-to-program synthesis}
\label{sec:program_synthesis}

\subsubsection{Code decoder}
\label{sec:code_decoder}

The code decoder $G_{\mathrm{code}}$ uses a second LoRA adapter over the same
Qwen2.5-Coder-1.5B base model. Its prompt contains the task instruction, the
query modality, the generated construction plan $\hat{r}$, the available
observation block $c_{\mathrm{obs}}$, and an output cue. The instruction
requires the generated script to define a top-level variable named
\texttt{part}, which is subsequently consumed by the execution pipeline.

The plan and code decoders therefore share a language-model backbone but use
different adapted weights. They are trained for different targets: the first
produces Construction IR, while the second produces executable Flluma Python
code.

\subsubsection{Deterministic repair and execution}
\label{sec:repair_and_exec}

Before execution, three narrow deterministic rewrites correct recurring
operation-alias and argument-format errors. These rules do not use execution
feedback, do not resample the program, and do not alter the predicted
construction beyond the predefined rewrite.

Each resulting program is then evaluated by the Flluma/OpenCASCADE kernel
using the five validity gates defined in
Section~\ref{sec:evaluation_axes}. Unless stated otherwise, the main
experiments use one generated program per query ($N=1$), with no
geometry-based test-time selection.

\subsection{Training}
\label{sec:stage3b}

Training is organised into four principal stages, with two short continuation
phases. Table~\ref{tab:training_stages} summarises which modules are updated at
each step.

\begin{table}[htbp]
  \centering
  \small
  \caption{Training schedule of MIRAGE-CAD. Stages 3b and 4b continue the
    preceding checkpoints rather than starting new models.}
  \label{tab:training_stages}

  \begin{tabular}{
      >{\raggedright\arraybackslash}p{0.07\linewidth}
      >{\raggedright\arraybackslash}p{0.27\linewidth}
      >{\raggedright\arraybackslash}p{0.38\linewidth}
      >{\raggedright\arraybackslash}p{0.14\linewidth}
    }
    \toprule
    Stage
     & Purpose
     & Trainable modules
     & Objective                                                    \\
    \midrule

    1
     & Representation alignment
     & Projection heads, point/STEP encoders, and $f_{\mathrm{ir}}$
     & $\mathcal{L}_{\mathrm{align}}$                               \\

    2
     & Latent prior fitting
     & Four modality-specific priors $\pi_m$
     & $\mathcal{L}_{\mathrm{prior}}$                               \\

    3
     & Plan-decoder adaptation
     & Prefix adapter $\Psi$ and LoRA-IR
     & Plan NLL                                                     \\

    3b
     & Cross-modal continuation
     & Prefix adapter $\Psi$ and LoRA-IR
     & Plan NLL                                                     \\

    4
     & Code-decoder training
     & LoRA-Code
     & Program NLL                                                  \\

    4b
     & Predicted-plan continuation
     & LoRA-Code
     & Program NLL                                                  \\

    \bottomrule
  \end{tabular}
\end{table}

Stage~1 learns the common construction-aligned representation. Text and image
backbones remain frozen, while the projection heads, point-cloud encoder, STEP
descriptor encoder, and IR encoder are trained with
$\mathcal{L}_{\mathrm{align}}$. Stage~2 freezes this representation and fits
the four modality priors independently.

Stage~3 jointly trains the prefix adapter and plan-decoder LoRA weights. It
starts from STEP-prior latents. Stage~3b resumes the same checkpoint and
continues training with latents from all four modality priors so that one plan
decoder can be used across modalities.

Stage~4 trains the code-decoder adapter on reference construction plans.
Because inference uses predicted plans, Stage~4b continues training with a
mixture of $70\%$ reference plans and $30\%$ grammar-valid predicted plans.
This continuation addresses the train--inference exposure mismatch, following
the same general motivation as scheduled sampling
\cite{bengio2015scheduledsampling}.

The retrieval index used by the baseline experiments is built after
representation alignment. It updates no model parameters and is not part of
the deployed training or inference path.

\section{Experimental Protocol}
\label{sec:setup}

\subsection{Data and evaluation splits}
\label{sec:data}

Experiments use FllumaOne-100K
\cite{zhan2026fllumaonecodenativemultimodalcad}, a corpus of 100,000
procedurally generated CAD parts. Each sample contains an executable Flluma
program, its Construction IR, natural-language descriptions, rendered views,
a surface point cloud, and STEP/B-Rep geometry. Reference programs are
executed and validated by the Flluma/OpenCASCADE kernel during corpus
generation.

The main experiments use 25,000 training, 2,500 validation, and 2,500 test
samples per modality. Point clouds contain 2,048 stored surface points and are
subsampled to 1,024 for MIRAGE-CAD. Image experiments use one isometric view
from the eight rendered views available for each part. The retrieval index is
built from training samples only.

Three evaluation settings are used. The default \textbf{IID} setting uses the
2,500 held-out test samples per modality. The \textbf{family-held-out} setting
removes four complete template families from training and evaluates on 2,923
samples from those families. Every operation type occurring in the held-out
families remains represented in the retained training data, so the split
primarily tests new combinations of known operations. Because this split also
uses a smaller training set, it is treated as a structural diagnostic rather
than a clean single-factor generalisation test.

The \textbf{cross-source} setting contains 400 externally authored parts from
the Fusion 360 Gallery
\cite{willis2021fusion360gallery}. All MIRAGE-CAD checkpoints used on this set
are trained only on FllumaOne.

\subsection{Experimental conditions}
\label{sec:baselines}

Table~\ref{tab:configurations} summarises the experimental conditions. These
rows should not be interpreted as a single-factor ablation table: the
controlled comparisons are defined pairwise, and different comparisons hold
different parts of the pipeline fixed.

\begin{table}[htbp]
  \centering
  \footnotesize
  \caption{Experimental conditions used in the main comparisons.}
  \label{tab:configurations}

  \begin{tabular}{
      >{\raggedright\arraybackslash}p{0.08\linewidth}
      >{\raggedright\arraybackslash}p{0.32\linewidth}
      >{\raggedright\arraybackslash}p{0.17\linewidth}
      >{\raggedright\arraybackslash}p{0.29\linewidth}
    }
    \toprule
    Arm
     & Construction conditioning
     & Explicit plan
     & Role                               \\
    \midrule

    \textbf{B0}
     & None
     & No
     & Direct query-to-code baseline      \\

    \textbf{B1a}
     & Continuous $\hat{z}_{\mathrm{ir}}$
     & No
     & One-epoch direct-latent arm        \\

    \textbf{B1b}
     & Continuous $\hat{z}_{\mathrm{ir}}$
     & No
     & Extended direct-latent arm         \\

    \textbf{B2}
     & Latent followed by plan
     & Generated
     & Deployed MIRAGE-CAD                \\

    \textbf{B2-Pred}
     & Latent followed by plan
     & Generated
     & One-epoch plan arm                 \\

    \textbf{B3}
     & Retrieved training construction
     & Retrieved
     & Retrieval reference                \\

    \textbf{B4}
     & Point cloud
     & No
     & Released CAD-Recode model          \\
    \bottomrule
  \end{tabular}
\end{table}

B0 removes the plan block together with the construction-conditioned pathway
that feeds it. It is therefore a pathway-level baseline, not an isolated
removal of textualisation.

B1 passes the predicted construction latent directly to the code decoder
through a prefix adapter, without generating a textual plan. B1a is the
one-epoch condition used for the closest comparison with B2-Pred. The two
conditions use the same one-epoch schedule and closely matched inference
exposure, with evaluated checkpoints at 3,000 and 3,125 updates,
respectively. Their conditioning interfaces remain different: B1a presents a
continuous latent, whereas B2-Pred presents generated text. B1b extends the
direct-latent training budget and is reported as an additional diagnostic.

B2 is the deployed MIRAGE-CAD pathway described in
Section~\ref{sec:method}. B3 replaces the generated plan with a plan retrieved
from the training index while retaining the same plan-to-code interface. B4 is
the released CAD-Recode checkpoint
\cite{rukhovich2025cadrecode}, used only for external point-cloud positioning.

Unless stated otherwise, generation is single-shot ($N=1$), with no
geometry-based candidate selection at test time.

\subsection{Evaluation metrics}
\label{sec:eval_protocols}

\textbf{Executable validity.}
Programs are evaluated with the five sequential gates defined in
Section~\ref{sec:evaluation_axes}: Python parsing, execution, solid building,
kernel validation, and STEP export. Each gate rate uses all attempted queries
as its denominator.

\textbf{Geometric fidelity.}
The internal evaluator samples the B-Rep surface directly and computes
symmetric Chamfer distance in mm$^2$ together with F-score at a threshold of
$1\%$ of the reference bounding-box diagonal (F@1\%). Both shapes are
represented by 1,024 surface samples. Geometry summaries are computed only
where the relevant metric is defined. Paired comparisons use the intersection
of metric-scoreable samples, while geometric coverage is reported separately
over all inputs.

Because F@1\% depends on finite surface sampling, we also report an empirical
self-sampling reference, denoted $F_{\mathrm{self}}$. It is 0.244 on the
internal test set and 0.304 on the cross-source set. These values are empirical
references rather than upper bounds.

Where an all-input F-score accounting is reported, non-scoreable outputs are
assigned zero for that accounting only; it is kept separate from conditional
per-part fidelity statistics.

\textbf{Parametric behaviour.}
Each exposed parameter is perturbed independently using

\begin{equation}
  \Delta =
  \{-25\%,\,-10\%,\,+10\%,\,+25\%\}.
\end{equation}

For the resulting perturbations we report $R_{\mathrm{build}}$, the fraction
for which the modified program remains five-gate valid, and
$R_{\mathrm{resp}}$, the fraction that additionally changes the geometry.
Parameter reachability is measured separately as the fraction of geometric
numeric arguments controlled through declared parameters rather than
hard-coded literals.

\textbf{Construction diagnostics.}
Agreement with the dataset reference construction is measured using four
per-sample quantities. Plan cosine compares embeddings of the generated and
reference Construction IR texts,

\begin{equation}
  \cos
  \left(
  E_{\mathrm{ir}}(\hat{r}),
  E_{\mathrm{ir}}(\mathbf{r})
  \right),
\end{equation}

whereas latent cosine compares
$\hat{z}_{\mathrm{ir}}$ with
$z_{\mathrm{ir}}=E_{\mathrm{ir}}(\mathbf{r})$. Op-Set F1 compares the sets of
\texttt{OP\_*} operations, and Op-Seq LCS compares their order. The
\texttt{PART} and \texttt{SEED} fields are replaced by fixed placeholders
before plan-level metrics are computed. These measures evaluate agreement with
one reference construction and are not treated as direct measures of geometry.

\subsection{Statistical analysis}
\label{sec:intervals}

Validity rates are reported with 95\% Wilson score intervals. When two
conditions are evaluated on the same sample ids, binary outcomes are compared
with exact McNemar tests. Paired per-part geometry comparisons use two-sided
sign tests unless another paired analysis is stated explicitly.

The diagnostic-value analysis is performed on the complete 500-sample
exposure-matched slice. AUROC measures how each construction diagnostic ranks
Build and STEP-export success, with 0.5 corresponding to chance ranking.
Spearman correlation is used for association with geometric fidelity.
Differences between diagnostics are estimated from 10,000 paired bootstrap
resamples. No classifier is fitted and no decision threshold is selected.

\subsection{External point-cloud positioning}
\label{sec:external_protocol}

The released CAD-Recode checkpoint
\cite{rukhovich2025cadrecode} and the MIRAGE-CAD point-cloud pathways are
evaluated on the same 400 Fusion 360 parts. This comparison uses a common
downstream evaluator, but the systems retain their released or deployed input
pipelines. CAD-Recode follows its released preprocessing, including
farthest-point downsampling to 256 points, whereas MIRAGE-CAD uses its
1,024-point normalised input. The experiment therefore compares complete
systems rather than isolating encoder architecture or point budget.

For this comparison, generated STEP files are re-tessellated and both
prediction and reference are normalised to a unit cube. Chamfer distance
follows the released CAD-Recode evaluation convention: 8,192 surface samples
per shape, bidirectional mean squared nearest-neighbour distances, summed and
multiplied by 1000. Volumetric IoU is computed from the tessellated meshes.
This external Chamfer measure is different from the internal mm$^2$ metric and
the two are never compared directly.

Coverage is reported over all 400 inputs. Fidelity comparisons use only parts
for which both systems provide the metric in question, with separate
denominators for Chamfer distance and IoU. Evaluator failures are reported
separately rather than silently discarded. Paired coverage is analysed with
exact McNemar tests; paired fidelity differences are accompanied by bootstrap
intervals and winner counts by exact sign tests.

The comparison is intended as system-level external positioning. It is not a
reproduction of CAD-Recode's published benchmark, because the systems differ
in training data, representation, preprocessing, and program interface.

\subsection{Implementation}
\label{sec:implementation}

Experiments were run on a single NVIDIA RTX 5060 Ti with 16 GB of memory.
The two language-model stages use Qwen2.5-Coder-1.5B
\cite{yang2025qwen25}, loaded with 4-bit NF4 weights and double quantisation
following QLoRA~\cite{dettmers2023qlora}. The plan and code decoders are
adapted with LoRA~\cite{hu2021lora}; their adapters are loaded sequentially at
inference to remain within the available memory.

Stages 1 and 2 train the smaller representation and prior networks without
4-bit quantisation. Optimiser settings, learning rates, sequence lengths,
batch sizes, and stage-specific training schedules are recorded with the
implementation details.

All trained conditions reported in the paper are single-seed runs. The paired
statistical tests therefore quantify variation across evaluated parts, not
variation across independent training seeds.

\section{Results}
\label{sec:results}

Unless stated otherwise, results are reported on the 2,500 held-out samples
per modality with single-shot decoding ($N=1$), after deterministic repair.
Comparisons on the same sample ids are analysed as paired observations.
Geometric comparisons use the subset on which both conditions provide the
relevant metric, while coverage is reported separately over all inputs as
defined in Section~\ref{sec:eval_protocols}.

\subsection{RQ1 --- What does construction mediation contribute?}
\label{sec:rq1}

We first compare the deployed construction-mediated pathway with direct
query-to-code generation on STEP input. Removing the construction pathway
reduces Build from 70.0\% to 35.4\% and STEP export from 66.2\% to 34.6\%
(Table~\ref{tab:main}). Syntax moves in the opposite direction, from 94.5\%
to 97.6\%. The improvement therefore cannot be explained by the mediated
model simply producing more parseable Python. Its main effect is on whether
the generated program reaches a buildable CAD model.

\begin{table}[htbp]
  \centering
  \footnotesize
  \caption{STEP results on the 2,500-sample held-out set. B0 removes the
    construction-mediated pathway, while B2 is deployed MIRAGE-CAD.}
  \label{tab:main}
  \begin{tabular}{llrrr}
    \toprule
    ID & Configuration                     & Syntax (\%)   & Build (\%)    & STEP Export (\%) \\
    \midrule
    B0 & Direct query $\rightarrow$ code
       & \textbf{97.6}                     & 35.4          & 34.6                             \\
    B2 & Generated plan $\rightarrow$ code
       & 94.5                              & \textbf{70.0} & \textbf{66.2}                    \\
    \bottomrule
  \end{tabular}
\end{table}

This comparison establishes the value of the construction-mediated pathway,
but does not isolate the textual plan. B0 removes the plan together with the
continuous construction representation that feeds it. We therefore introduce
B1, in which the predicted construction latent
$\hat{z}_{\mathrm{ir}}$ conditions the code decoder directly and no textual
plan is generated.

\subsubsection{Continuous versus textual construction conditioning}
\label{sec:direct_latent}

\begin{figure}[htbp]
  \centering
  \includegraphics[width=\textwidth]{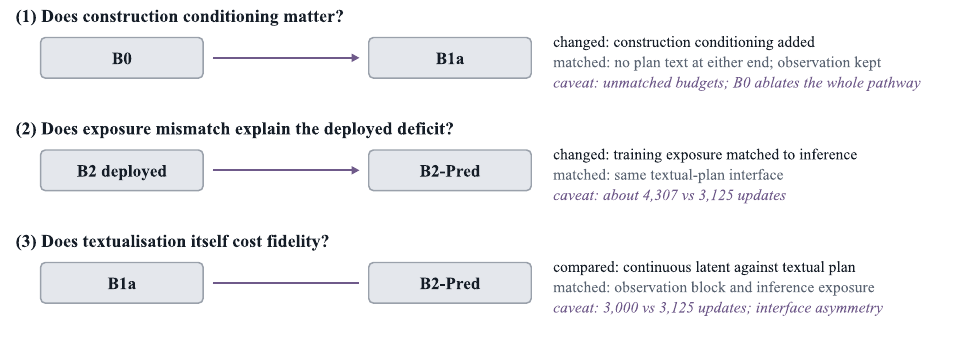}
  \caption{\textbf{Comparisons used to separate construction conditioning,
      training exposure, and textualisation.}
    B0--B1a tests construction conditioning without plan text.
    B2--B2-Pred tests the effect of matching code-decoder training to
    predicted-plan inference. B1a--B2-Pred gives the closest comparison
    between continuous and textual construction interfaces. These are not
    factorial ablations: training budgets differ in the first two comparisons,
    while B1a--B2-Pred retains a small checkpoint difference
    ($3{,}000$ versus $3{,}125$ updates) and an interface asymmetry.}
  \label{fig:arms}
\end{figure}

Figure~\ref{fig:arms} shows how the three comparisons separate construction
conditioning, training exposure, and textualisation.

The controlled runs use the same first $500$ STEP samples from the held-out
split, with greedy decoding, $N=1$, and no repair. B1a is evaluated at its
best validation checkpoint at step $3{,}000$, while B2-Pred uses its final
checkpoint at step $3{,}125$. Both follow the same one-epoch schedule and have
closely matched inference exposure, but their optimisation budgets are not
identical.

\begin{table}[htbp]
  \centering
  \footnotesize
  \caption{Continuous construction conditioning and textual plan conditioning
    on the same 500 STEP samples. B1a and B2-Pred form the closest comparison,
    evaluated at 3,000 and 3,125 updates, respectively; B2 is the deployed
    curriculum reference at approximately 4,307 updates.}
  \label{tab:direct_latent}

  \begin{tabular}{
      >{\raggedright\arraybackslash}p{0.30\linewidth}
      >{\centering\arraybackslash}p{0.18\linewidth}
      >{\centering\arraybackslash}p{0.20\linewidth}
      >{\centering\arraybackslash}p{0.18\linewidth}
    }
    \toprule
    Metric
     & B1a latent
     & \textbf{B2-Pred plan}
     & B2 deployed                                  \\
    \midrule
    Syntax (\%)
     & 95.8                  & 95.8          & 95.2 \\
    Build (\%)
     & 86.2                  & \textbf{90.4} & 71.4 \\
    STEP export (\%)
     & 85.6                  & \textbf{90.0} & 67.4 \\
    Unconditional F@1 / $F_{\mathrm{self}}$ (\%)
     & 73.0                  & \textbf{76.8} & 53.6 \\
    Bbox within $\pm10\%$ (\%)
     & 94.4                  & \textbf{97.1} & 90.6 \\
    \bottomrule
  \end{tabular}
\end{table}

The first result is that explicit text is not required for strong
reconstruction. Direct conditioning on the continuous construction latent
reaches 86.2\% Build and 85.6\% STEP export. A longer three-epoch B1 run
reaches 93.6\% on both gates. Construction conditioning can therefore support
strong reconstruction without an intermediate textual plan.

The second result concerns the deployed training curriculum. B2-Pred trains
the code decoder on the predicted plans it will receive at inference. Build
rises from 71.4\% for deployed B2 to 90.4\%, with 108 discordant pairs
favouring B2-Pred against 13 ($p=8.4\times10^{-20}$). STEP export rises from
67.4\% to 90.0\% ($126:13$, $p=2.1\times10^{-24}$). B2-Pred also improves
paired geometric fidelity on the common scoreable subset. The comparison is
not update-matched---B2 received roughly 4,307 updates and B2-Pred 3,125---but
the result shows that training-condition mismatch is a major source of the
deployed plan pathway's deficit.

The closest comparison of textualisation itself is therefore B1a against
B2-Pred. Both arms provide a geometric score on 393 parts. The paired sign
test does not establish a fidelity direction ($182:211$, $p=0.158$). For
Chamfer distance, the estimated B2-Pred--B1a difference is
$-0.0126$~mm$^2$, with a 95\% bootstrap interval of
$[-0.0347,0.0051]$. For F@1 the difference is $0.0040$, with interval
$[-0.0008,0.0089]$. Both intervals lie inside the empirically derived
practical-effect margins used in this comparison. On this STEP sample and for
this single-seed checkpoint pair, we therefore detect no practically material
per-part fidelity cost from expressing the construction representation as
text. This is deliberately narrower than claiming general equivalence.

B2-Pred has 4.2 percentage points higher Build than B1a
(90.4\% versus 86.2\%, $p=0.031$). We treat this difference cautiously. The
comparison is single-seed, and B1a must additionally learn a continuous
adapter into the code decoder, whereas text already uses the decoder's native
interface. The experiment therefore does not establish a stable coverage
advantage for textualisation.

\subsubsection{Diagnostic value of the explicit plan}
\label{sec:plan_diagnostic}

The direct-latent comparison shows that textualising the construction
representation is not required for strong reconstruction. We therefore ask a
different question: whether agreement at the explicit plan interface is
informative about what happens downstream. We evaluate this on the same
500-sample B2-Pred condition.

\begin{figure}[htbp]
  \centering
  \includegraphics[width=\textwidth]{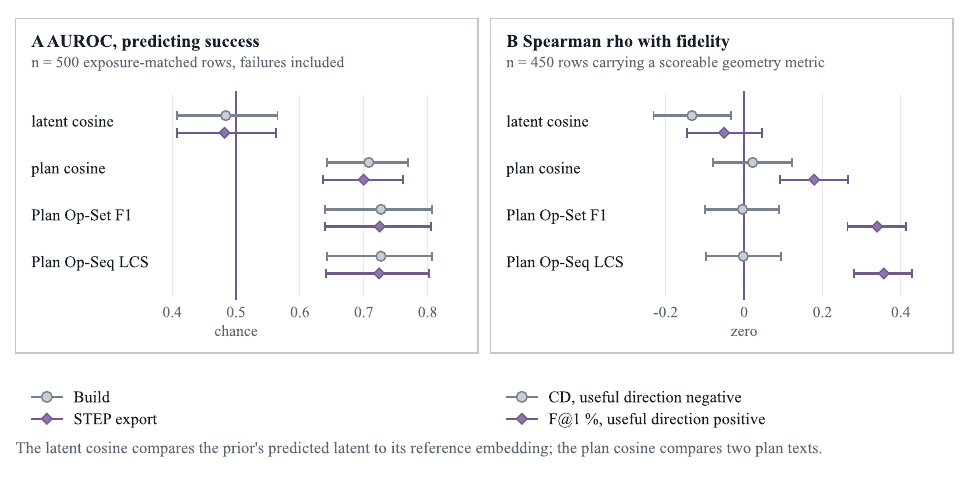}
  \caption{\textbf{Diagnostic value of agreement at the construction-plan
      interface.}
    Panel A reports AUROC for Build and STEP-export success on the
    $500$ exposure-matched samples. Panel B reports Spearman correlations
    with Chamfer distance and F@1 on the $450$ samples with scoreable geometry.
    Negative correlation is favourable for Chamfer distance and positive
    correlation for F@1. Whiskers show 95\% paired bootstrap intervals.
    Plan-based diagnostics require the reference construction and are therefore
    available only for offline evaluation, not as deployment-time confidence
    scores.}
  \label{fig:diagnostic_value}
\end{figure}

\begin{table}[htbp]
  \centering
  \footnotesize
  \caption{Diagnostic value of plan-level agreement. AUROC is computed on all
    500 exposure-matched STEP rows; geometric correlations use the 450 rows with
    both CD and F@1 scores.}
  \label{tab:plan_diagnostic}
  \begin{tabular}{lrrrr}
    \toprule
    Diagnostic
                    & Build AUROC $\uparrow$
                    & Export AUROC $\uparrow$
                    & $\rho$ with CD
                    & $\rho$ with F@1                                                            \\
    \midrule
    Latent cosine   & 0.484                   & 0.482             & $\mathbf{-0.133}$ & $-0.051$ \\
    Plan cosine     & 0.708                   & 0.700             & $+0.022$          & $+0.179$ \\
    Plan Op-Set F1  & \textbf{0.727}          & \textbf{0.725}
                    & $-0.004$                & $\mathbf{+0.340}$                                \\
    Plan Op-Seq LCS & \textbf{0.727}          & 0.724
                    & $-0.002$                & $\mathbf{+0.357}$                                \\
    \bottomrule
  \end{tabular}
\end{table}

Plan-level agreement is informative about downstream validity. Op-Set F1 and
Op-Seq LCS both reach about 0.73 AUROC for Build and STEP export, whereas the
scalar latent cosine is about 0.48. Relative to latent cosine, the three plan
diagnostics improve AUROC by approximately 0.22--0.24, with the paired
bootstrap intervals excluding zero.

This result concerns one scalar summary of the latent,
$\cos(\hat{z}_{\mathrm{ir}},z_{\mathrm{ir}})$; it does not imply that the full
latent vector contains no failure information. It also remains a supervised
diagnostic because plan cosine, Op-Set F1 and Op-Seq LCS require the dataset's
reference construction.

The association with geometry depends on the metric. Op-Set F1 and Op-Seq LCS
correlate with scale-normalised F@1
($\rho=0.340$ and $0.357$), but are essentially unrelated to raw Chamfer
distance. Plan metrics also compare against one reference construction, so a
low score does not by itself imply an incorrect final shape. Their practical
value here is that the intermediate construction can be inspected and scored
separately from the final program.

\subsection{RQ2 --- What information do the construction and observation channels carry?}
\label{sec:ablation_prefix}

We next test whether the plan decoder actually uses its continuous
construction prefix. On 500 STEP queries, replacing the correct prefix with
another sample's prefix reduces plan cosine from 0.886 to 0.047 and Op-Set F1
from 88.0\% to 34.6\%. The conditioning vector therefore carries
sample-specific construction information.

The effect is largely invisible to Build. The correct and shuffled prefixes
reach 71.4\% and 68.6\% Build, respectively ($p=0.37$), although median
Chamfer distance rises from 2.84 to 22.11~mm$^2$. A corrupted construction
signal can therefore still produce an executable solid while substantially
changing its geometry. This is a direct example of why executable validity and
geometric fidelity must be reported separately.

\begin{figure}[htbp]
  \centering
  \includegraphics[width=\textwidth]{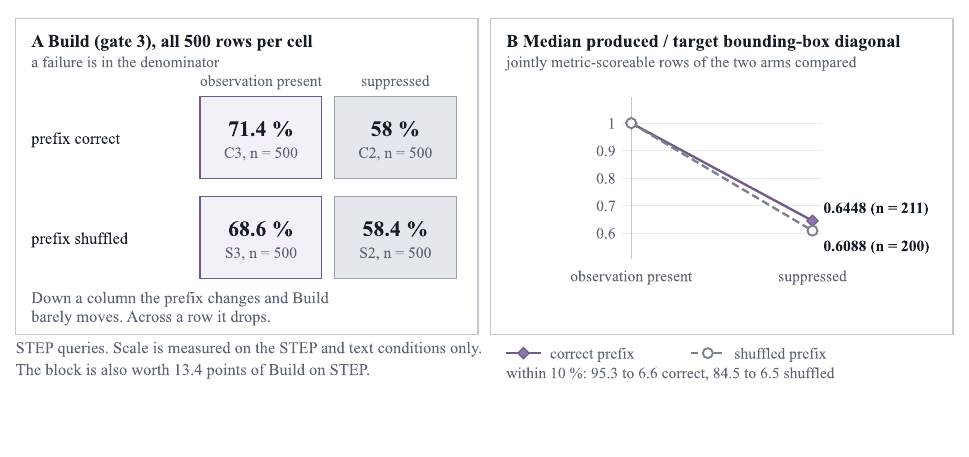}
  \caption{\textbf{Prefix and observation interventions on STEP queries.}
    Panel A reports Build for the four combinations of correct or shuffled
    construction prefix and present or suppressed observation. Each condition
    includes all $500$ queries. Panel B shows the median produced-to-target
    bounding-box diagonal ratio before and after observation suppression for
    the correct- and shuffled-prefix conditions. The two comparisons use their
    own jointly scoreable subsets ($n=211$ and $n=200$). Observation is
    suppressed only at inference; the decoders were trained with the
    observation block present.}
  \label{fig:obs_intervention}
\end{figure}

The construction prefix is not the only input to the decoders. We therefore
suppress the query-derived observation block at both decoder inputs, using the
same trained checkpoints and the same $500$ held-out rows.
Figure~\ref{fig:obs_intervention} shows the resulting effects on Build and
absolute scale.

On STEP, suppressing the observation block reduces Build from 71.4\% to
58.0\% ($p=2.0\times10^{-9}$). On text, Build changes from 59.6\% to
58.0\% and the difference is not detected ($p=0.57$). Geometry gives a
different picture: median Chamfer rises from 2.60 to 53.26~mm$^2$ on STEP and
from 4.26 to 59.42~mm$^2$ on text.

The strongest effect is on absolute scale. For STEP, the median
produced-to-target bounding-box ratio falls from 1.000 to 0.645 when the
observation is suppressed, and the fraction within $\pm10\%$ of the target
size falls from 95.3\% to 6.6\%. For text, the ratio falls from 1.000 to
0.622 and the corresponding within-$10\%$ rate from 87.6\% to 10.1\%.
With a shuffled construction prefix but the observation still present, the
STEP ratio remains 1.000 and 84.5\% of the paired parts remain within 10\% of
the correct size.

These interventions show that the deployed STEP and text pathways rely on the
observation channel to preserve absolute metric scale. This statement is
specific to the tested decoders: suppression is performed at inference on
models trained with the block present and does not estimate the performance of
a decoder retrained without it.

The construction signal remains relevant when this auxiliary channel is
closed. With the observation suppressed, correct and shuffled prefixes give
58.0\% and 58.4\% Build ($p=0.95$), while their geometry still differs. The
continuous construction representation and the observation block therefore
carry different information: the former carries construction-relevant
information, whereas the latter supplies additional query evidence, including
absolute metric scale for modalities in which such information is available.

\subsection{RQ3 --- Can one shared pipeline serve four modalities?}
\label{sec:generation_results}

One shared plan decoder and one shared code decoder produce executable programs
from all four input modalities. On $2{,}500$ held-out queries per modality, the
generated pathway reaches $55.4$--$70.0\%$ Build and $52.3$--$66.2\%$ STEP
export.

\begin{table}[htbp]
  \centering
  \footnotesize
  \caption{Generated-path results on the four held-out modalities,
    $2{,}500$ queries each.}
  \label{tab:generation}
  \begin{tabular}{lrrrr}
    \toprule
    Modality
     & Syntax (\%)
     & Prog-Op-F1 (\%)
     & Build (\%)
     & STEP Export (\%) \\
    \midrule
    Text
     & 98.3
     & 80.1
     & 57.2
     & 53.1             \\
    Image
     & 94.3
     & 79.1
     & 57.8
     & 52.3             \\
    Point cloud
     & 89.5
     & 73.9
     & 55.4
     & 52.6             \\
    STEP/B-Rep
     & 94.5
     & 85.1
     & \textbf{70.0}
     & \textbf{66.2}    \\
    \bottomrule
  \end{tabular}
\end{table}

The table supports a multimodal system claim rather than a ranking of the
modalities. The encoders differ in capacity, downstream training begins from
STEP-conditioned data, and the observation blocks carry different information.
The observation intervention makes this last point concrete: the deployed STEP
pathway leads text by $11.8$ Build points, whereas both reach $58.0\%$ Build
after the observation block is suppressed.

A supporting training experiment also shows that the shared plan decoder is
sensitive to the latent distribution used during training. Continuing Stage~3
from STEP latents to a four-modality mixture raises average Build from
$42.9\%$ to $60.1\%$, with gains of $18.9$--$28.7$ percentage points for the
three non-STEP modalities, while STEP decreases by $4.4$ points. The complete
per-modality training analysis is reported in Appendix~\ref{app:stage3b}.

\subsection{RQ4 --- What do retrieval and external evaluation reveal?}
\label{sec:rq4}

\subsubsection{Construction redundancy and the coverage--fidelity trade-off}
\label{sec:redundancy}

Retrieval is unusually strong on the procedural corpus. On the IID split the
retrieval variants reach 95.0--99.5\% Build. Holding out four complete
template families does not remove this advantage: retrieval remains at
97.6--100.0\% Build, while generated-path Build ranges from 45.0\% to 62.4\%.
The reason is visible in construction space. The nearest retrieved training
plan has a median operation-set F1 of exactly 1.0 against the query's reference
plan. A typical held-out query therefore has another training part with the
same operation set even when its template family has been excluded. The
family-held-out split is consequently a useful structural diagnostic, but not
a construction-held-out benchmark.

Geometry changes the interpretation of retrieval. For STEP queries on the
2,923-part family-held-out split, the generated and retrieved arms are jointly
F@1-scoreable on 1,330 parts. On this common subset, the generated arm reaches
a median F@1 of 0.162, or 66.2\% of $F_{\mathrm{self}}$, compared with 0.111
(45.6\%) for retrieval; generation wins 982 paired comparisons to 347
($p=1.7\times10^{-70}$).

When all 2,923 parts are included in the separate all-input accounting and a
non-scoreable output contributes zero, the ordering reverses: retrieval has a
mean F@1 of 0.099 (40.6\% of $F_{\mathrm{self}}$), compared with 0.066
(26.9\%) for generation. Retrieval therefore gives much greater coverage,
whereas generation is more faithful among the parts both pathways can score.
The two summaries answer different questions and are reported together.

The cross-source results are consistent with the observation-channel
intervention. On externally authored parts, the generated STEP pathway has a
median produced-to-target bounding-box ratio of 0.993, compared with 0.575 for
its retrieval counterpart. The point-cloud pathway does not retain absolute
metric scale, and does not show the same advantage.

\subsubsection{External positioning against CAD-Recode}
\label{sec:external_positioning}

We next compare the MIRAGE point-cloud pathways with the released CAD-Recode
checkpoint~\cite{rukhovich2025cadrecode} on the same 400 Fusion 360 parts,
using the common evaluator defined in Section~\ref{sec:external_protocol}.
This is a system-level comparison: the models retain their own point-cloud
preprocessing and training distributions.

\begin{table}[htbp]
  \centering
  \footnotesize
  \caption{External point-cloud positioning on 400 Fusion 360 parts.
    Coverage is reported as MIRAGE versus CAD-Recode over all inputs.
    CD and IoU values are paired medians on the subsets scoreable by both
    systems for the corresponding metric.}
  \label{tab:external_summary}

  \begin{tabular}{
      >{\raggedright\arraybackslash}p{0.18\linewidth}
      >{\centering\arraybackslash}p{0.24\linewidth}
      >{\centering\arraybackslash}p{0.25\linewidth}
      >{\centering\arraybackslash}p{0.25\linewidth}
    }
    \toprule
    MIRAGE pathway
     & Coverage
     & Paired CD
     & Paired IoU                         \\
    \midrule

    Generated plan
     & 226/400 vs 392/400
     & 55.78 / 0.117 \newline ($n=222$)
     & 0.0604 / 0.9226 \newline ($n=200$) \\

    Prior-NN-IR
     & 392/400 vs 392/400
     & 17.59 / 0.187 \newline ($n=385$)
     & 0.1621 / 0.9426 \newline ($n=369$) \\

    \bottomrule
  \end{tabular}
\end{table}

The autonomous MIRAGE generated-plan pathway produces measurable geometry for
226 of 400 parts, compared with 392 for CAD-Recode. The gap remains large after
conditioning on common success: on 222 jointly CD-scoreable parts the paired
medians are 55.78 for MIRAGE and 0.117 for CAD-Recode; on the 200 jointly
IoU-scoreable parts they are 0.0604 and 0.9226.

Retrieval recovers coverage but not fidelity. Prior-NN-IR and CAD-Recode each
produce measurable geometry for 392 of 400 parts, with exactly seven
discordant successes in each direction ($p=1$). Yet among the 369 parts with
an IoU value from both systems, the paired median is 0.1621 for Prior-NN-IR
and 0.9426 for CAD-Recode. The same ordering appears in Chamfer distance.

This comparison places a clear boundary on the present system. The current
MIRAGE point-cloud pathway is substantially weaker than a specialised
point-cloud-to-CAD model on externally authored geometric reconstruction.
Retrieval can reproduce its coverage, but not its geometric accuracy. These
results do not test whether construction should be represented continuously
or textually; they show the current limit of the multimodal pathway on a
specialised reconstruction task.

\begin{figure}[htbp]
  \centering
  \includegraphics[width=\textwidth]{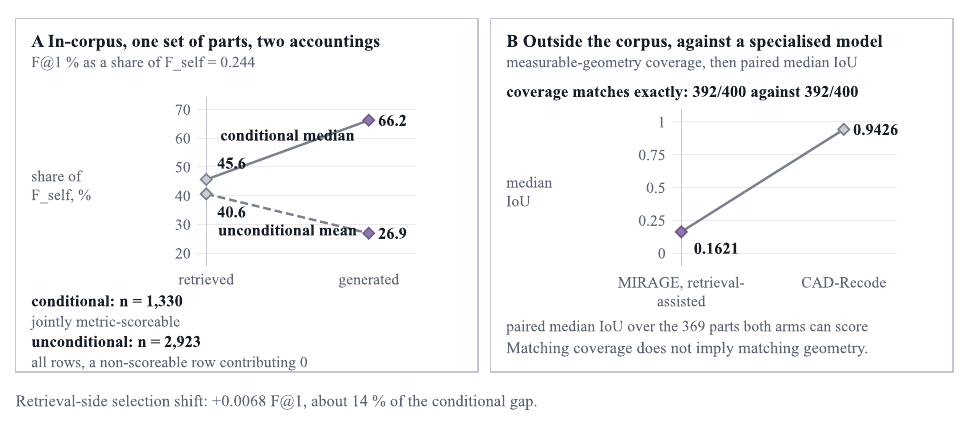}
  \caption{\textbf{Coverage and fidelity answer different questions.}
    Panel A compares the generated and retrieved STEP pathways on the
    family-held-out split. Among the $1{,}330$ parts scoreable by both arms,
    generation has the higher conditional median F@1. When all $2{,}923$
    parts are included and non-scoreable outputs are assigned zero, retrieval
    has the higher mean. Panel B compares retrieval-assisted MIRAGE with
    CAD-Recode on Fusion 360 parts. Both produce measurable geometry for
    $392/400$ inputs, but CAD-Recode has a much higher paired median IoU on
    the $369$ parts scoreable by both systems.}
  \label{fig:coverage_fidelity}
\end{figure}

Figure~\ref{fig:coverage_fidelity} places the internal and external results
side by side. In both settings, coverage and conditional geometric fidelity
lead to different conclusions.

\subsection{RQ5 --- Are the generated programs parametrically responsive?}
\label{sec:editability_results}

Finally, we evaluate parametric behaviour on the $74$ of $100$ generated STEP
programs that pass all five validity gates. Each declared parameter is perturbed
independently by $\pm10\%$ and $\pm25\%$, giving $1{,}608$ perturbations in
total.

\begin{table}[htbp]
  \centering
  \footnotesize
  \caption{Parametric behaviour of the generated STEP pathway.}
  \label{tab:parametricity}
  \begin{tabular}{rrrrr}
    \toprule
    Parts & Perturbations & Rebuild-valid (\%) & Responsive (\%) & Reachability (\%) \\
    \midrule
    74    & 1,608         & 99.8               & 82.2            & 29.3              \\
    \bottomrule
  \end{tabular}
\end{table}

Of the $1{,}608$ tested perturbations, $1{,}604$ ($99.8\%$) remain five-gate
valid after the edit, and $1{,}321$ ($82.2\%$) produce a detectable geometric
change. Most declared parameters can therefore be perturbed without breaking
the generated program, although some edits leave the geometry unchanged.

Parameter reachability is more limited. The mean per-part reachability is
$29.3\%$, meaning that, on average, fewer than one third of a program's
geometric numeric arguments are controlled through declared parameters.
The main limitation is therefore not the stability of parameter edits, but the
fraction of the generated geometry that is exposed through those parameters.

\section{Discussion and Limitations}
\label{sec:discussion}

\subsection{Construction representation and textual mediation}
\label{sec:disc_mediation}

The experiments separate two questions that are easily conflated: whether a
construction representation is useful for CAD generation, and whether that
representation needs to be expressed as text.

The first question is answered clearly. Removing the construction-mediated
pathway causes a large loss in Build and STEP export, while a code decoder
conditioned directly on the continuous construction latent reconstructs
strongly without generating a textual plan. Construction conditioning is
therefore useful, but the reconstruction capability cannot be attributed to
textualisation itself.

The exposure-matched comparison sharpens this point. On the tested STEP
checkpoint pair, continuous and textual construction conditioning show no
detected practically material difference in per-part geometric fidelity. The
comparison is single-seed and the evaluated checkpoints differ slightly in
update count, so this is not a general equivalence result. It does, however,
rule out the stronger interpretation that an explicit textual plan is required
for accurate reconstruction.

The role of the plan is better understood as an interface. It is readable,
can be scored independently of the final program, and provides an intermediate
point at which the generation process can be inspected. Its agreement with the
reference construction is also informative about downstream execution:
plan-level diagnostics reach AUROC values around 0.70--0.73 for Build and STEP
export, whereas the scalar latent cosine is close to chance. These are
supervised evaluation diagnostics, since they require the dataset reference
construction, rather than confidence scores available to the deployed system.

The architecture also contains a second information path. For STEP and text
queries, inference-time suppression of the query-derived observation block
strongly disrupts absolute scale, whereas changing the construction prefix has
a different effect on geometry. The deployed system therefore uses the
continuous construction representation and the observation block for different
purposes. This conclusion is specific to the tested intervention: the decoder
was trained with the observation block present, so the experiment does not
estimate what a model retrained without that channel could learn.

\subsection{Validity, fidelity, and parametric behaviour}
\label{sec:disc_evaluation}

The experiments also show why executable validity should be treated as a gate
rather than as a measure of reconstruction quality.

Replacing a query's construction prefix with that of another part produces a
large deterioration in plan agreement and geometric fidelity while leaving
Build statistically unchanged. The external comparison provides an independent
example: the retrieval-assisted MIRAGE pathway and CAD-Recode produce
measurable geometry for the same number of Fusion 360 parts, yet their paired
IoU values remain widely separated. Similar coverage therefore does not imply
similar geometry.

Parametric behaviour introduces a third distinction. Almost all tested edits
to exposed parameters rebuild successfully, and most produce a geometric
response, but the mean per-part parameter reachability is only $29.3\%$.
High responsiveness of the parameters that exist therefore does not imply
broad parameterisation of the generated program.

Validity, geometric fidelity, and parametric behaviour are thus empirically
dissociable and should be reported separately. A validity rate answers whether
the program runs; geometric metrics answer whether it reconstructs the target;
and perturbation tests answer how the generated program behaves when edited.
None of these quantities can substitute for the others.

\subsection{Non-identifiability of construction histories}
\label{sec:non_identifiable}

A final CAD solid does not uniquely determine the sequence of operations that
produced it. The same geometry may be obtained through different sketches,
Boolean sequences, or feature orders. Recovering the original construction
history from geometry alone is therefore not a well-defined objective.

For this reason, the Construction IR stored in the dataset is treated as a
\emph{reference construction}, not as a unique ground-truth history.
Plan-level metrics measure agreement with that reference and remain useful for
controlled comparisons, but they may penalise an alternative construction that
produces equally valid geometry. Their absolute values should therefore not be
interpreted as construction accuracy.

A more complete evaluation would compare a prediction against a set of
construction histories consistent with the target geometry rather than against
a single reference. Constructing such a set remains an open problem.

\subsection{Procedural redundancy and benchmark design}
\label{sec:benchmark_redundancy}

The retrieval experiments reveal a second property of the evaluation setting.
Holding out complete template families does not remove construction-level
redundancy: retrieval remains extremely strong, and the nearest retrieved plan
has a median operation-set F1 of 1.0 against the query's reference plan.

Family membership is therefore a poor proxy for construction novelty in this
procedural corpus. Parts from different template families can still share the
same operation skeleton. A benchmark intended to measure compositional
generalisation should consequently define separation in construction space,
rather than relying only on generator-defined family labels.

This observation is related to the broader effects of training-set redundancy
and nearest-neighbour reuse reported in prior work
\cite{lee2022dedup,khandelwal2020knnlm}. For CAD generation, a simple
training-set retrieval baseline is particularly informative because it exposes
redundancy directly at the level of construction.

\subsection{Limitations}
\label{sec:discussion_limitations}

Several limitations bound the conclusions of this study.

First, the comparison between continuous and textual construction interfaces
uses one trained checkpoint per condition. The paired analysis measures
variation across test parts, not training-seed variation, and the two evaluated
checkpoints differ slightly in optimisation budget. The small Build advantage
of the textual arm should therefore not be interpreted as an established
general effect.

Second, external point-cloud reconstruction remains a clear weakness. On 400
Fusion 360 parts, the released CAD-Recode checkpoint
\cite{rukhovich2025cadrecode} substantially outperforms both MIRAGE
point-cloud pathways in geometric fidelity. The comparison does not isolate an
architectural cause because the systems differ in training data, point
representation, encoder, and code interface. It nevertheless establishes a
practical boundary: the current MIRAGE point-cloud pathway is not competitive
with a specialised point-cloud-to-CAD reconstruction model.

Third, the training corpus is procedurally generated and contains no recorded
human modelling sessions. Its construction histories are plausible rather than
observed, and the measured redundancy limits what can be inferred from
in-distribution and family-held-out results.

Fourth, the four modality encoders differ substantially in capacity, so their
results compare the implemented pathways rather than the intrinsic
informativeness of the modalities. In addition, the current point-cloud
preprocessing removes absolute scale before encoding. This is a limitation of
the implementation, not of point clouds as a representation.

Finally, the demonstrated parametric responsiveness applies to a limited
portion of each generated program: fewer than one third of geometric numeric
arguments are reachable through declared parameters. Assemblies, mating
constraints, and full sketch-constraint graphs are also outside the present
single-part scope. Additional implementation diagnostics and known defects are reported in
Appendix~\ref{app:defects}.

\section{Conclusion}
\label{sec:conclusion}

MIRAGE-CAD studies construction-mediated multimodal generation of executable
CAD programs through a shared construction representation. From text, rendered images, point clouds, and
STEP/B-Rep observations, the system generates executable Python CAD programs
that can be run by an OpenCASCADE-based kernel and exported as STEP. Across
2,500 held-out queries per modality, the generated pathway reaches
55.4--70.0\% Build and 52.3--66.2\% STEP export without retrieval at
inference.

The experiments refine the role of the intermediate construction plan.
Construction conditioning is important, but strong reconstruction does not
require the construction representation to be textualised. Under the tested
exposure-matched STEP conditions, the textual interface shows no detected
practically material per-part fidelity cost relative to direct continuous
conditioning. Its measured value instead lies in providing an explicit,
separately scorable interface through which construction can be inspected and
analysed.

The results also show that executable validity, geometric fidelity, and
parametric behaviour describe different properties of a generated CAD
program. A program may execute successfully while reconstructing the wrong
geometry, and responsive exposed parameters may still control only a small
fraction of its geometric numeric content. These properties should therefore
be evaluated separately.

Finally, retrieval exposes substantial construction redundancy in the
procedural corpus, while external comparison with CAD-Recode
\cite{rukhovich2025cadrecode} identifies a clear limitation of the current
point-cloud pathway. Future work should improve construction-space separation,
increase parameter reachability, and extend the framework beyond single-part
solids while developing evaluation methods that account for the
non-identifiability of valid CAD construction histories.

\section*{Code and data availability}
\label{sec:availability}

The training and evaluation code, Construction IR tooling, and scripts used
to generate and validate the quantitative results reported in this paper are
publicly available at \url{\repourl}.

\emph{Corpus.}
The experiments use the 100,000-model kernel-validated corpus described in
Section~\ref{sec:data}. The corpus is distributed separately under its
original terms and is not redistributed as part of the MIRAGE-CAD repository.

\emph{CAD-Recode.}
The external comparison in Section~\ref{sec:external_positioning} uses the
released CAD-Recode checkpoint under the CC BY-NC 4.0 licence. The checkpoint
is loaded from the upstream release at run time and is not redistributed in
the MIRAGE-CAD repository. The accompanying evaluation scripts record the
checkpoint identity and preprocessing configuration required to reproduce the
comparison.

\appendix
\counterwithin{table}{section}
\counterwithin{figure}{section}
\renewcommand{\thetable}{\Alph{section}.\arabic{table}}
\renewcommand{\thefigure}{\Alph{section}.\arabic{figure}}
\section{Construction IR Grammar and Operation Inventory}
\label{app:ir_grammar}

Section~\ref{sec:formulation_ir} introduces the Construction IR at a conceptual level.
This appendix gives the complete statement grammar and operation vocabulary used in the
experiments, together with their frequencies in the training split.

\subsection{Statement grammar}
\label{app:ir_statements}

A Construction IR document is a newline-separated sequence composed of four statement
types. Across the $25{,}000$ STEP-modality training samples, the corpus contains
$25{,}000$ \texttt{PART} statements, $175{,}141$ \texttt{PARAM} statements,
$137{,}018$ \texttt{F} statements, and $25{,}000$ \texttt{END} statements.

\begin{center}
  \footnotesize
  \begin{tabular}{
      >{\raggedright\arraybackslash}p{0.11\linewidth}
      >{\raggedright\arraybackslash}p{0.74\linewidth}
    }
    \toprule
    Statement & Syntax and role                                                       \\
    \midrule
    \texttt{PART}
              & \texttt{PART <name>}. Opens the document and identifies the part.     \\

    \texttt{PARAM}
              & \texttt{PARAM <name> <value> MIN <lo> MAX <hi> SEM <semantic>}.
    Declares a named numeric parameter and its admissible range. Feature
    arguments may reference the parameter as \texttt{@<name>}; such
    references are used when measuring parameter reachability in
    \S\ref{sec:editability_results}.                                                  \\

    \texttt{F}
              & \texttt{F <id> OP\_<TYPE> SEM <semantic> ROLE <role> [DEP <id>]
                    [TARGET <ref>] PARAMS <k>=<v> \ldots}. Describes one construction
    feature. \texttt{DEP} records a dependency on an earlier feature,
    while \texttt{TARGET} may refer to a sub-entity of a previous feature,
    for example \texttt{extrude\_001.top\_face}.                                      \\

    \texttt{END}
              & Terminates the document.                                              \\
    \bottomrule
  \end{tabular}
\end{center}

\subsection{Semantic and role vocabularies}
\label{app:ir_metrics}

In addition to the operation token, feature records contain semantic and role annotations.
The \texttt{SEM} field takes \textbf{118} distinct values in the training split. Frequent
examples include \texttt{hole\_radius}, \texttt{thickness}, \texttt{hole\_spacing},
\texttt{placement}, and \texttt{flange}. The field is therefore treated as an open
semantic vocabulary rather than a fixed categorical set.

The \texttt{ROLE} field is restricted to seven values:

\begin{center}
  \footnotesize
  \texttt{primary\_body} \quad \texttt{secondary\_feature} \quad
  \texttt{functional\_feature} \quad \texttt{functional\_cut} \\
  \texttt{pattern\_feature} \quad \texttt{profile\_definition} \quad
  \texttt{finishing\_feature}
\end{center}

Neither field is used by the operation-level metrics. Op-Set F1 and Op-Seq LCS are
computed only from \texttt{OP\_*} identifiers. By contrast, \texttt{SEM} and
\texttt{ROLE} remain part of the normalised IR text and can therefore affect the full-IR
embedding used for plan cosine.

\subsection{Operation vocabulary}
\label{app:ir_vocabulary}

The operation vocabulary contains $|\mathcal{V}|=44$ tokens. The complete inventory is
reported in Table~\ref{tab:ir_vocabulary}.

\emph{Documents} denotes the number of training samples in which an operation occurs at
least once, and \emph{document coverage} reports the same quantity as a percentage of the
$25{,}000$ samples. Since one sample may contain several operation types, document
coverage does not sum to $100$~\%. \emph{Total occurrences} counts all \texttt{F}
statements associated with each operation and sums to $137{,}018$.

\begin{table}[htbp]
  \centering
  \scriptsize
  \caption{Complete Construction IR operation vocabulary over the $25{,}000$
    STEP-modality training samples. \emph{Documents} counts samples containing the
    operation at least once; \emph{document coverage} is the corresponding percentage of
    the training split. \emph{Total occurrences} counts all associated \texttt{F}
    statements.}
  \label{tab:ir_vocabulary}
  \begin{tabular}{lrrr}
    \toprule
    Operation                           & Documents & Document coverage (\%) & Total occurrences \\
    \midrule
    \texttt{OP\_HOLE\_PATTERN}          & 22{,}094  & 88.38                  & 24{,}303          \\
    \texttt{OP\_TRANSLATE}              & 9{,}623   & 38.49                  & 15{,}545          \\
    \texttt{OP\_BOOLEAN\_JOIN}          & 11{,}667  & 46.67                  & 15{,}065          \\
    \texttt{OP\_SKETCH}                 & 11{,}254  & 45.02                  & 13{,}287          \\
    \texttt{OP\_BOX}                    & 8{,}255   & 33.02                  & 9{,}637           \\
    \texttt{OP\_PLATE}                  & 7{,}905   & 31.62                  & 7{,}905           \\
    \texttt{OP\_CHAMFER}                & 7{,}233   & 28.93                  & 7{,}233           \\
    \texttt{OP\_EXTRUDE}                & 6{,}735   & 26.94                  & 6{,}735           \\
    \texttt{OP\_RECTANGULAR\_POCKET}    & 6{,}598   & 26.39                  & 6{,}598           \\
    \texttt{OP\_CYLINDER}               & 5{,}104   & 20.42                  & 6{,}180           \\
    \texttt{OP\_FILLET}                 & 3{,}699   & 14.80                  & 3{,}699           \\
    \texttt{OP\_SKETCH\_ON\_FACE}       & 1{,}761   & 7.04                   & 3{,}522           \\
    \texttt{OP\_REVOLVE}                & 2{,}044   & 8.18                   & 2{,}044           \\
    \texttt{OP\_LOFT}                   & 2{,}033   & 8.13                   & 2{,}033           \\
    \texttt{OP\_FACE\_EXTRUDE\_ADD}     & 1{,}761   & 7.04                   & 1{,}761           \\
    \texttt{OP\_FACE\_EXTRUDE\_CUT}     & 1{,}761   & 7.04                   & 1{,}761           \\
    \texttt{OP\_SWEEP\_TUBE}            & 1{,}730   & 6.92                   & 1{,}730           \\
    \texttt{OP\_MIRROR}                 & 1{,}251   & 5.00                   & 1{,}589           \\
    \texttt{OP\_LINEAR\_PATTERN}        & 873       & 3.49                   & 873               \\
    \texttt{OP\_SHELL}                  & 543       & 2.17                   & 543               \\
    \texttt{OP\_CIRCULAR\_PATTERN}      & 538       & 2.15                   & 538               \\
    \texttt{OP\_BOOLEAN\_CUT}           & 491       & 1.96                   & 491               \\
    \texttt{OP\_THREADED\_HOLE}         & 467       & 1.87                   & 467               \\
    \texttt{OP\_PROFILE\_CUT}           & 442       & 1.77                   & 442               \\
    \texttt{OP\_BOOLEAN\_INTERSECT}     & 383       & 1.53                   & 383               \\
    \texttt{OP\_STANDOFF\_ARRAY\_PLATE} & 268       & 1.07                   & 268               \\
    \texttt{OP\_BOSSED\_PLATE}          & 242       & 0.97                   & 242               \\
    \texttt{OP\_LOFT\_ADAPTER}          & 230       & 0.92                   & 230               \\
    \texttt{OP\_SENSOR\_MOUNT\_PLATE}   & 227       & 0.91                   & 227               \\
    \texttt{OP\_REVOLVED\_BUSHING}      & 213       & 0.85                   & 213               \\
    \texttt{OP\_RAIL\_MOUNT\_PLATE}     & 153       & 0.61                   & 153               \\
    \texttt{OP\_MIRRORED\_LUG\_PLATE}   & 131       & 0.52                   & 131               \\
    \texttt{OP\_PIN\_CLUSTER}           & 129       & 0.52                   & 129               \\
    \texttt{OP\_TORUS}                  & 125       & 0.50                   & 125               \\
    \texttt{OP\_RIBBED\_SUPPORT\_PLATE} & 122       & 0.49                   & 122               \\
    \texttt{OP\_CLAMP\_PAD\_PLATE}      & 118       & 0.47                   & 118               \\
    \texttt{OP\_ELLIPSOID}              & 116       & 0.46                   & 116               \\
    \texttt{OP\_PRISM}                  & 116       & 0.46                   & 116               \\
    \texttt{OP\_SHELLED\_COVER\_PLATE}  & 114       & 0.46                   & 114               \\
    \texttt{OP\_WEDGE}                  & 86        & 0.34                   & 86                \\
    \texttt{OP\_CAPSULE}                & 71        & 0.28                   & 71                \\
    \texttt{OP\_CLIPPED\_OFFSET\_STOCK} & 71        & 0.28                   & 71                \\
    \texttt{OP\_CONE}                   & 65        & 0.26                   & 65                \\
    \texttt{OP\_SPHERE}                 & 57        & 0.23                   & 57                \\
    \bottomrule
  \end{tabular}
\end{table}

\subsection{Operation-frequency distribution}
\label{app:ir_distribution}

The vocabulary is strongly imbalanced. \texttt{OP\_HOLE\_PATTERN} appears in
$88.4$~\% of training samples, whereas 23 of the 44 operations occur in fewer than
$2$~\% of samples. This long tail provides the context for the rare-operation analysis reported
in Appendix~\ref{app:anchor}.

The imbalance also affects the interpretation of plan-level metrics. An aggregate Op-Set
F1 score is influenced primarily by frequent operations and can remain high even when
rare operations are poorly represented. Op-Set F1 and Op-Seq LCS are therefore useful as
relative diagnostics across conditions, but they should not be read as uniform measures of
performance over the 44 operation classes.

\section{STEP/B-Rep Descriptor Definition and Effective Input}
\label{app:step_features}

Section~\ref{sec:encoders} describes the STEP/B-Rep branch at the architectural level.
This appendix gives the four descriptor streams used by the implementation and clarifies
which of them contain sample-specific information in the corpus used for the experiments.

\subsection{Descriptor streams}
\label{app:step_streams}

The STEP/B-Rep encoder accepts four inputs:

\begin{itemize}
  \item a \textbf{global descriptor} containing $50$ kernel-derived statistics;
  \item a \textbf{face stream} of shape $64\times28$ with a $64$-element mask, where each
        face is represented by $11$ surface-type indicators and $17$ continuous fields;
  \item an \textbf{edge stream} of shape $128\times24$ with a $128$-element mask, where each
        edge is represented by $9$ curve-type indicators and $15$ continuous fields;
  \item a \textbf{relation stream} in $\mathbb{R}^{32}$.
\end{itemize}

\subsection{Effective input in the present corpus}
\label{app:step_effective}

For the corpus used in this study, the extractor produces no populated per-entity face or
edge records: both local streams are fully masked, and the relation stream is constant
across samples. Sample-specific variation in the STEP input is therefore carried by the
$50$-dimensional global descriptor. Accordingly, the STEP results reported in this paper
should be interpreted as results obtained primarily from kernel-derived global statistics,
not from a learned representation of detailed B-Rep topology.

This distinction limits what can be concluded from the STEP branch. The experiments do not
evaluate whether explicit face-, edge-, or topology-aware encodings would improve the
results, nor do they provide evidence against such encodings. The architecture supports
those streams, but the current extractor does not exercise them with independent
sample-specific content.

We also avoid treating the deployed branch as formally equivalent to a
$50$-dimensional multilayer perceptron. Establishing such an equivalence would require
showing that the masked or constant streams make no contribution anywhere in the forward
pass, including through biases and normalisation terms. That property was not verified and
is not needed for the interpretation above.

\subsection{Global descriptor layout}
\label{app:step_layout}

Table~\ref{tab:step_descriptor} lists the $50$ global features in extractor order. Slot
indices are zero-based. The first slot is the kernel validity flag; all remaining entries
are transformed using $\log(1+x)$.

\begin{table}[htbp]
  \centering
  \footnotesize
  \caption{Layout of the $50$-dimensional global STEP/B-Rep descriptor.
    Slot indices are zero-based and follow the extractor order. All entries
    except the validity flag are transformed using $\log(1+x)$.}
  \label{tab:step_descriptor}

  \begin{tabular}{
      >{\raggedright\arraybackslash}p{0.11\linewidth}
      >{\centering\arraybackslash}p{0.08\linewidth}
      >{\raggedright\arraybackslash}p{0.67\linewidth}
    }
    \toprule
    Slots & Count                                                              & Content \\
    \midrule

    0
          & 1
          & \textbf{Validity}: \texttt{brep\_valid}, represented as 0 or 1.              \\

    1--6
          & 6
          & \textbf{Topology counts}: numbers of solids, shells, faces, wires,
    edges, and vertices.                                                                 \\

    7--9
          & 3
          & \textbf{Bounding box}: extents along the three coordinate axes.              \\

    10
          & 1
          & \textbf{Surface area}: total B-Rep surface area.                             \\

    11
          & 1
          & \textbf{Volume}: solid volume.                                               \\

    12--22
          & 11
          & \textbf{Surface-type counts}: \texttt{plane}, \texttt{cylinder},
    \texttt{cone}, \texttt{sphere}, \texttt{torus},
    \texttt{bezier\_surface}, \texttt{bspline\_surface},
    \texttt{surface\_of\_revolution},
    \texttt{surface\_of\_extrusion},
    \texttt{offset\_surface}, and \texttt{other\_surface}.                               \\

    23--31
          & 9
          & \textbf{Curve-type counts}: \texttt{line}, \texttt{circle},
    \texttt{ellipse}, \texttt{hyperbola}, \texttt{parabola},
    \texttt{bezier\_curve}, \texttt{bspline\_curve},
    \texttt{offset\_curve}, and \texttt{other\_curve}.                                   \\

    32--35
          & 4
          & \textbf{Edge--face valence}: counts of edges with valence
    1, 2, 3, and 4 or more.                                                              \\

    36--41
          & 6
          & \textbf{Incidence and manifoldness}: face--wire, face--edge, and
    face--edge-adjacency incidence counts, followed by boundary,
    manifold, and non-manifold edge counts.                                              \\

    42--45
          & 4
          & \textbf{Face-area statistics}: minimum, maximum, mean, and sum.              \\

    46--49
          & 4
          & \textbf{Edge-length statistics}: minimum, maximum, mean, and sum.            \\

    \midrule
    \textbf{Total}
          & \textbf{50}
          &                                                                              \\
    \bottomrule
  \end{tabular}
\end{table}

\subsection{Absolute-scale information}
\label{app:step_scale}

The global descriptor retains absolute metric quantities. Slots $7$--$9$ contain the three
bounding-box extents, while slots $10$ and $11$ contain surface area and volume. These
features provide an explicit source of absolute-scale information to the STEP pathway and
are consistent with the scale-preservation behaviour observed in the observation-channel
intervention.

The point-cloud pathway differs because its input is divided by the maximum radius before
encoding, removing absolute size before the network sees the cloud. This is a property of
the current preprocessing rather than an intrinsic limitation of point clouds. Text can
retain absolute dimensions when they are stated explicitly in the query, whereas the image
observation block is empty. These differences should be kept in mind when comparing scale
behaviour across modalities.

\section{Additional Training and Representation Diagnostics}
\label{app:diagnostics}

\subsection{Direct-latent training sensitivity and practical-effect analysis}
\label{app:budget}

Section~\ref{sec:direct_latent} compares B1a and B2-Pred at $3{,}000$ and $3{,}125$
updates. The budgets are close but not identical. This subsection reports the effect of
additional optimisation on the direct-latent arm and gives the practical-effect analysis used
for the one-epoch comparison.

\paragraph{Training budget}
Increasing B1 from $3{,}000$ to $9{,}000$ updates raises Build from $86.2$~\% to
$93.6$~\% and increases the number of exported parts from $428$ to $468$. On the $398$
parts jointly scoreable at both checkpoints, median Chamfer changes from $2.293$ to
$2.222$~mm$^2$ ($p = 0.0058$). The additional optimisation therefore has a much larger
effect on coverage than on per-part geometric fidelity.

The longer run also changes syntax and output diversity. Syntax increases from $95.8$~\%
at one epoch to $100.0$~\% at three epochs. Validation loss decreases from $0.0943$ to
$0.0828$, while the number of distinct API-call skeletons decreases from $75$ to $53$.
Thus, the shorter checkpoint is not simply an earlier copy of the same solution: it has a
higher validation loss and produces a more diverse set of program skeletons.

Both direct-latent checkpoints remain stronger than deployed B2 on the comparisons for which
they were evaluated. For the three-epoch checkpoint, Build has $128$ discordant pairs in
favour of B1 and $17$ in favour of B2 ($p = 3.0\times10^{-22}$); STEP export gives
$148:17$ ($p = 2.9\times10^{-27}$). On the $312$ parts jointly scoreable for geometry, B1
is closer on $201$ parts, with median Chamfer $2.212$ versus $2.541$~mm$^2$
($p = 3.9\times10^{-7}$). Because each paired comparison has its own jointly scoreable
subset, these conditional medians are interpreted only within the corresponding pair.

\paragraph{Practical-effect analysis for B1a and B2-Pred}
The one-epoch comparison contains $393$ parts with a geometric score from both arms. The
paired sign test does not establish a direction ($182:211$, $p = 0.158$). The effect
estimates and bootstrap intervals are:

\begin{center}
  \footnotesize
  \begin{tabular}{lrrr}
    \toprule
    Quantity
     & Estimate
     & 95~\% bootstrap CI
     & Margin              \\
    \midrule
    $\Delta$ Chamfer, paired median (mm$^2$)
     & $-0.0126$
     & $[-0.0347,+0.0051]$
     & $0.055$             \\
    $\Delta$ F@1, paired mean
     & $+0.0040$
     & $[-0.0008,+0.0089]$
     & $0.012$             \\
    \bottomrule
  \end{tabular}
\end{center}

The margins are empirical rather than pre-registered. They are taken from B1a's paired gain
over deployed B2 on the corresponding metric and rows: $0.055$~mm$^2$ for Chamfer and
$0.012$ for F@1. Both intervals include zero and remain inside these margins. Under this
criterion, no practically material per-part difference is detected for this checkpoint pair.
This is not treated as general equivalence between continuous and textual construction
representations. The analysis is single-seed, and the bootstrap resamples held-out parts;
it therefore does not estimate training-seed variance.

\paragraph{Coverage difference}
B2-Pred exceeds B1a by $4.2$ percentage points of Build ($54:33$, $p = 0.031$) and
$4.4$ points of STEP export ($56:34$, $p = 0.026$), while syntax is identical at $17:17$.
The unconditional F@1 difference is $+0.009$, with interval $[+0.001,+0.018]$. Since the
paired per-part fidelity comparison does not establish a difference, this all-input advantage
is mainly associated with the larger number of successful outputs.

Two experimental asymmetries should be kept in view. B1a is restored from its
best-by-validation checkpoint, whereas B2-Pred is evaluated at its final step. B1a's best
and final checkpoints differ by $125$ updates and $0.0008$ in validation loss. B2-Pred's
validation loss rises monotonically because its validation split contains reference plans while
training moves toward predicted-plan exposure, so the same checkpoint-selection rule would not
measure the intended condition. In addition, B1a must learn a continuous adapter into the code
decoder's embedding space, whereas B2-Pred uses the decoder's native textual interface. These
differences are small relative to the main reconstruction gains but matter when interpreting the
$4.2$-point coverage gap. No seed-variance estimate is available for this comparison.

\subsection{Plan-exposure and latent-distribution training diagnostics}
\label{app:exposure}

The two continuation stages address different train--inference mismatches. Stage~3b changes
the distribution of construction latents seen by the plan decoder, while Stage~4b changes the
distribution of plans seen by the code decoder.

\subsubsection{Stage 3b: cross-modal continuation}
\label{app:stage3b}

Stage~3 trains the plan decoder on STEP-prior latents. Stage~3b resumes that checkpoint and
continues training with latents from all four modality priors.

\begin{table}[htbp]
  \centering
  \footnotesize
  \caption{Effect of Stage~3b on Build, with $2{,}500$ held-out rows per modality and
    95~\% Wilson intervals. The $\Delta$ column is computed from unrounded values.}
  \label{tab:stage3b}
  \begin{tabular}{lccr}
    \toprule
    Modality    & before (\%)       & after (\%)        & $\Delta$         \\
    \midrule
    STEP        & 74.3 [72.6, 76.0] & 70.0 [68.1, 71.7] & \textbf{$-4.4$}  \\
    Point cloud & 29.9 [28.1, 31.7] & 55.4 [53.5, 57.4] & \textbf{$+25.6$} \\
    Text        & 28.5 [26.7, 30.3] & 57.2 [55.3, 59.1] & \textbf{$+28.7$} \\
    Image       & 38.8 [36.9, 40.8] & 57.8 [55.8, 59.7] & \textbf{$+18.9$} \\
    \midrule
    Average     & 42.9              & 60.1              & \textbf{$+17.2$} \\
    \bottomrule
  \end{tabular}
\end{table}

The four-modality continuation raises average Build from $42.9$~\% to $60.1$~\%. The three
non-STEP modalities gain between $18.9$ and $28.7$ percentage points, while STEP decreases
by $4.4$ points. The result indicates that the shared plan decoder is sensitive to the latent
distribution used during training. It does not establish that the modalities contain equivalent
information: the encoders differ in capacity and the observation channels differ in content.

\subsubsection{Stage 4b: predicted-plan continuation}
\label{app:stage4b}

Stage~4 trains the code decoder primarily with reference plans, whereas deployment supplies
predicted plans. Stage~4b introduces a $70/30$ mixture of reference and predicted plans. The
following diagnostic uses Variant~C STEP queries with $n=100$ per row.

\begin{center}
  \footnotesize
  \begin{tabular}{l>{\raggedright\arraybackslash}p{0.24\linewidth}rrr}
    \toprule
    Inference input & Training                                    & Syntax (\%) & Prog-Op-F1 (\%) & Build (\%)  \\
    \midrule
    Reference plan  & Stage 4 only                                & 100.0       & 97.9            & 95          \\
                    & Stage 4b: 70~\% reference + 30~\% predicted & 100.0       & 97.6            & 89          \\
    Predicted plan  & Stage 4 only                                & 99.0        & 76.5            & 28          \\
                    & Stage 4b: 70~\% reference + 30~\% predicted & 99.0        & 84.1            & \textbf{67} \\
    \bottomrule
  \end{tabular}
\end{center}

With the Stage~4 checkpoint, Build falls from $95$~\% under a reference plan to $28$~\%
under a predicted plan. Stage~4b raises the predicted-plan result to $67$~\%, while Build
with a reference plan decreases from $95$~\% to $89$~\%. Training entirely on predicted
plans, as in B2-Pred, reaches $90.4$~\% Build on the 500-row controlled evaluation. These
results show that plan exposure is a major source of the deployed code decoder's performance
gap. The $70/30$ continuation reduces the mismatch but does not remove it.

\subsection{Construction-anchor analysis}
\label{app:anchor}

Six oversampled operation families occupy $1.8$--$7.0$~\% of the training rows and show
failure rates of $80$--$92$~\% across modalities. The same lack of separation is already
visible in the reference Construction-IR embeddings: pairwise cosine similarity reaches
$0.878$ for sweep-tube instances and $0.9996$ for circular-pattern instances, compared with
$0.006$ for a random contrast set.

Because these values are computed before the modality encoders enter the alignment, the
under-separation is present at or before the shared construction anchor. Encoder-specific
errors may still add to it, but they cannot fully explain the pattern. The analysis therefore
identifies the anchor representation as one source of difficulty for these operation families;
it does not identify which part of the anchor pipeline---text serialisation, truncation,
encoder capacity, contrastive objective, batch composition, or class imbalance---is responsible.

\subsection{Runtime failures and deterministic repair}
\label{app:failures}

\paragraph{Failure taxonomy}
Each failed program is assigned to one of five categories by matching the runtime error string
in a fixed order. Percentages below are relative to failures within each modality.

\begin{center}
  \scriptsize
  \setlength{\tabcolsep}{3.5pt}

  \begin{tabular}{
      >{\raggedright\arraybackslash}p{0.12\linewidth}
      >{\centering\arraybackslash}p{0.10\linewidth}
      >{\centering\arraybackslash}p{0.13\linewidth}
      >{\centering\arraybackslash}p{0.14\linewidth}
      >{\centering\arraybackslash}p{0.09\linewidth}
      >{\centering\arraybackslash}p{0.13\linewidth}
      >{\centering\arraybackslash}p{0.09\linewidth}
    }
    \toprule
    Modality
     & Failure rate (\%)
     & Keyword mismatch
     & Attribute hallucination
     & Syntax
     & Topology reference
     & Other                   \\
    \midrule

    STEP/B-Rep
     & 30.0
     & 30.8
     & 13.8
     & 18.2
     & 7.2
     & 30.0                    \\

    Point cloud
     & 44.6
     & 21.5
     & 25.1
     & 23.5
     & 8.7
     & 21.2                    \\

    Text
     & 42.8
     & 36.1
     & 23.4
     & 4.0
     & 12.8
     & 23.7                    \\

    Image
     & 42.2
     & 28.4
     & 18.8
     & 13.5
     & 12.0
     & 27.2                    \\
    \bottomrule
  \end{tabular}
\end{center}

Keyword mismatches and attribute hallucinations are name-level errors and are the cases most
directly addressed by deterministic rewriting. The \emph{Other} category accounts for
$21.2$--$30.0$~\% of failures and includes semantic or kernel-level errors such as a correct
API name used with the wrong type, value, arity, or shape, and solids that fail to export.

\paragraph{Deterministic repair}
Three narrow rewrites are applied before execution. At the plan level, one rule normalises a
recurring alias in the face-extrude family when the record belongs to a face-sketch chain. At
the program level, two rules correct recurring keyword-format errors:
\texttt{extrude\_on\_face} expects \texttt{sketch=} where the model often emits
\texttt{profile=}, and \texttt{profile\_cut\_on\_face} expects a three-component offset where
the model sometimes emits two.

Across the four $2{,}500$-row modality evaluations, these rules change Build by $+0.28$,
$+0.60$, $0.00$, and $+0.16$ percentage points, converting $26$ failures into successes
among $10{,}000$ rows. Their effect is therefore small relative to the remaining name-level
and semantic failures.

The procedure is deterministic and does not form an execution-feedback loop. Programs are not
re-executed after observing an error, runtime messages are not passed back to the model, and no
attempt is made to revise the predicted construction. The plan-level rule applies only to
generated plans; the two program-level rewrites are applied uniformly to all generated
candidates.

\section{Observation-Channel and Prefix Interventions}
\label{app:interventions}

This appendix gives the full intervention design underlying
Section~\ref{sec:ablation_prefix}. All conditions use the same Stage~3b and Stage~4b
checkpoints, the same $500$ held-out rows, and greedy decoding. Only the construction prefix
or the query-derived observation block is changed at inference. When the observation is
suppressed, it is removed from both decoder inputs so that the generated plan cannot retain
information from a block that is absent only at the code stage.

\subsection{Full crossed intervention}
\label{app:crossed}

\paragraph{Prefix intervention}
The prefix experiment compares five conditioning sources. A shuffled prefix is taken from a
different sample, so the decoder receives a valid continuous prefix carrying the wrong
construction information.

\begin{table}[htbp]
  \centering
  \footnotesize
  \caption{Prefix intervention on $500$ paired STEP queries. \emph{Zero} supplies $K$ zero
    vectors with $\Psi$ bypassed; \emph{constant} is $\Psi(\mathbf{0})$; \emph{shuffled} uses
    another sample's prefix; \emph{deployed} uses the predicted latent from $\pi_m$; and
    \emph{oracle} encodes the reference plan before applying $\Psi$. Build is reported with a
    95~\% Wilson interval and was not evaluated for the two synthetic-prefix conditions.}
  \label{tab:ablation_prefix}
  \begin{tabular}{lrrrr}
    \toprule
    Prefix source     & plan cos       & Op-Set F1 (\%) & Op-Seq LCS (\%) & Build (\%)               \\
    \midrule
    zero              & 0.057          & 38.1           & 28.3            & ---                      \\
    constant          & 0.082          & 35.9           & 26.8            & ---                      \\
    shuffled          & 0.047          & 34.6           & 26.0            & 68.6 [64.4, 72.5]        \\
    \textbf{deployed} & \textbf{0.886} & \textbf{88.0}  & \textbf{80.1}   & 71.4 [67.3, 75.2]        \\
    \emph{oracle}     & \emph{0.898}   & \emph{88.7}    & \emph{79.8}     & \emph{69.2 [65.0, 73.1]} \\
    \bottomrule
  \end{tabular}
\end{table}

Replacing the deployed prefix with a shuffled one reduces plan cosine from $0.886$ to
$0.047$ and Op-Set F1 from $88.0$~\% to $34.6$~\%. The shuffled condition is also below the
zero-prefix condition in Op-Set F1 ($34.6$ versus $38.1$~\%), which is consistent with the
decoder responding to the content of the prefix rather than only to its presence.

Using the reference-plan latent produces plan cosine $0.898$ and Op-Set F1 $88.7$~\%, close
to the deployed prior's $0.886$ and $88.0$~\%. The paired test does not detect a difference
($p = 0.26$). This suggests that, in this setting, the prior is not the dominant source of
plan-quality error; it is not an equivalence result.

\paragraph{Observation intervention}

\begin{table}[htbp]
  \centering
  \footnotesize
  \caption{Observation intervention on the same 500 held-out rows.
    Suppression is applied at both decoder inputs. Build is reported with a
    95\% Wilson interval. C1 and C0 are inference-time suppression conditions
    and should not be interpreted as retrained architectures.}
  \label{tab:obs_bypass}

  \begin{tabular}{llcrr}
    \toprule
    Condition
     & Plan
     & Obs.
     & STEP Build (\%)
     & Text Build (\%)            \\
    \midrule

    C3 deployed
     & Yes
     & Yes
     & \textbf{71.4} [67.3, 75.2]
     & \textbf{59.6} [55.2, 63.8] \\

    C2 plan-only
     & Yes
     & No
     & \textbf{58.0} [53.6, 62.2]
     & \textbf{58.0} [53.6, 62.2] \\

    C1 observation-only
     & No
     & Yes
     & 0.0
     & 0.2                        \\

    C0 neither
     & No
     & No
     & 0.0
     & 0.0                        \\

    \bottomrule
  \end{tabular}
\end{table}

For STEP queries, removing the observation block reduces Build from $71.4$~\% to
$58.0$~\% ($97:30$ discordant pairs, $p = 2.0\times10^{-9}$). For text, Build changes from
$59.6$~\% to $58.0$~\% ($79:71$, $p = 0.57$). Geometry is affected in both modalities:
median Chamfer rises from $2.60$ to $53.26$~mm$^2$ on STEP ($204$ of $211$ pairs favour
the observation-present condition, $p = 2.1\times10^{-51}$) and from $4.26$ to
$59.42$~mm$^2$ on text ($151$ of $169$, $p = 2.3\times10^{-27}$).

C1 and C0 are not estimates of a separately trained observation-only or unconditional
architecture. They are inference-time suppressions applied to decoders trained with the full
input structure. The trained plan-free baseline B0, for example, reaches $35.4$~\% Build with
the observation present, whereas removing the plan from the deployed decoder at inference
reduces Build to $0.0$--$0.2$~\%. The difference reflects distribution shift as well as the
information removed. In C0, all $500$ prompts contain no query information and are
byte-identical; only two distinct programs are emitted across the set.

\paragraph{Crossing prefix and observation interventions}
The two interventions can be combined on the same rows using the same shuffle permutation:

\begin{center}
  \footnotesize
  \begin{tabular}{
      l
      >{\raggedright\arraybackslash}p{0.24\linewidth}
      >{\raggedright\arraybackslash}p{0.24\linewidth}
    }
    \toprule
     & Observation present
     & Observation suppressed \\
    \midrule
    Plan correct
     & 71.4\%
     & 58.0\%                 \\
    Plan shuffled
     & 68.6\%
     & \textbf{58.4\%}        \\
    \bottomrule
  \end{tabular}
\end{center}

With the observation present, shuffling the prefix changes Build from $71.4$~\% to
$68.6$~\% ($111:97$, $p = 0.37$). With the observation suppressed, the corresponding
values are $58.0$~\% and $58.4$~\% ($115:117$, $p = 0.95$). The interaction is also not
detected ($p = 0.21$, paired sign test on the per-row difference of differences). Build is
therefore relatively insensitive to the corrupted construction signal in both observation
conditions, even though geometry changes substantially. In the crossed suppressed cell,
median Chamfer is $49.0$ versus $98.8$~mm$^2$ ($p = 0.002$).

That geometric comparison uses only $116$ jointly scoreable rows. The two suppressed
conditions individually yield scoreable surfaces for $238$ and $241$ parts, but their
intersection is much smaller. The result is therefore reported together with its denominator
rather than generalised to the full 500-row set.

\subsection{Absolute-scale analysis}
\label{app:scale}

The strongest and most consistent effect of the observation intervention is on absolute size.
Each pair below uses the rows on which both conditions provide the relevant geometric score.

\begin{center}
  \footnotesize
  \begin{tabular}{
      l
      >{\raggedright\arraybackslash}p{0.20\linewidth}
      >{\raggedright\arraybackslash}p{0.16\linewidth}
      r
    }
    \toprule
     & Median produced/target bbox diagonal
     & Within $\pm10\%$
     & Paired $n$                           \\
    \midrule

    STEP, observation present
     & 1.000
     & 95.3\%
     & \multirow{2}{*}{211}                 \\

    STEP, observation suppressed
     & 0.645
     & 6.6\%
     &                                      \\

    STEP shuffled, observation present
     & 1.000
     & 84.5\%
     & \multirow{2}{*}{200}                 \\

    STEP shuffled, observation suppressed
     & 0.609
     & 6.5\%
     &                                      \\

    Text, observation present
     & 1.000
     & 87.6\%
     & \multirow{2}{*}{169}                 \\

    Text, observation suppressed
     & 0.622
     & 10.1\%
     &                                      \\

    \bottomrule
  \end{tabular}
\end{center}

With the observation block present, the median bounding-box ratio is $1.000$ in all three
comparisons and $84.5$--$95.3$~\% of parts lie within $10$~\% of the target size. After
suppression, the median ratio falls to $0.609$--$0.645$ and the within-$10$~\% rate to
$6.5$--$10.1$~\%. The same direction is observed for STEP with either the correct or the
shuffled prefix and for text. For the tested deployed decoders, the observation channel is
therefore the dominant source of absolute metric-scale information in these conditions.

This intervention does not imply that a decoder trained without the observation block could
not learn scale through another route. Nor does it assign construction structure to the
textual plan. The direct-latent experiment shows that strong reconstruction is possible
without plan text, while the observation intervention isolates a separate dependence on
query-derived metric evidence.

The cross-source STEP result provides a complementary, but separate, observation. On the
400 external parts, the generated STEP pathway has a median produced-to-target bounding-box
ratio of $0.993$, compared with $0.575$ for retrieval. That comparison is correlational and
uses a different dataset; it is not another estimate of the intervention above.

\paragraph{STEP and text under matched observation conditions}
Both modalities use the same $500$ sample ids:

\begin{center}
  \footnotesize
  \begin{tabular}{lrrrrr}
    \toprule
                           & STEP & text & $\Delta$         & discordant       & $p$                \\
    \midrule
    deployed               & 71.4 & 59.6 & +11.8 pp         & 109 : 50         & $3.3\times10^{-6}$ \\
    observation suppressed & 58.0 & 58.0 & \textbf{+0.0 pp} & \textbf{47 : 47} & \textbf{1}         \\
    \bottomrule
  \end{tabular}
\end{center}

STEP leads text by $11.8$ Build points in the deployed condition. After suppressing the
observation block, the two Build rates are both $58.0$~\%. This equality is specific to
Build. Geometry deteriorates strongly in both modalities, so the matched gate rate should not
be read as matched reconstruction fidelity.

Chamfer distance and F@1 also respond differently to the scale perturbation. On the $151$
parts jointly scoreable for the plan-only and shuffled-with-observation conditions, median CD
is $49.28$ versus $22.08$, whereas median F@1 is $0.051$ versus $0.027$. Chamfer is sensitive
to absolute scale, while the F@1 threshold is normalised by the reference bounding-box
diagonal. The opposite ordering illustrates why the two geometric measures are reported
separately.

\subsection{Image-path null control}
\label{app:null_control}

Image queries provide no observation block at either decoder input. Suppressing that block
should therefore leave the prompt unchanged. This gives a direct null control for the
intervention implementation.

For all $500$ evaluated image samples, C3 and C2 produce byte-identical programs; the same is
true for C1 and C0. Program identity was verified using SHA-256 hashes. By contrast, the
corresponding STEP and text comparisons are identical for $0$ of $500$ samples. The null
control confirms that the suppression operation is inert when the observation field is empty.

\subsection{Scope of the intervention}
\label{app:intervention_scope}

All conditions in this appendix are inference-time interventions on decoders trained with the
observation block present. They measure the dependence of the deployed checkpoints on that
channel, not the achievable performance of architectures retrained without it. Suppression
therefore introduces a distribution shift, particularly in C1 and C0.

Geometric comparisons are also conditional on joint metric scoreability. Their denominators
are consequently smaller than the corresponding all-input Build denominators, and the crossed
suppression cell is especially selective. Coverage and conditional fidelity are reported
separately for this reason.

\section{Retrieval Redundancy and Benchmark Diagnostics}
\label{app:redundancy}

Section~\ref{sec:redundancy} shows that holding out complete template families does not remove the advantage of retrieval. This appendix reports the full family-held-out results, quantifies construction-skeleton redundancy, and examines the effect of conditioning geometric comparisons on joint scoreability.

\subsection{Full family-held-out results}
\label{app:family_heldout}

Four template families are excluded from training, and evaluation is performed on the $2{,}923$ samples belonging to those families. All training stages are rerun on this partition with the same hyperparameters as in the main split.

\begin{table}[htbp]
  \centering
  \footnotesize
  \caption{Family-held-out structural diagnostic on the $2{,}923$ held-out-family samples. Variant A is Direct-NN-IR, B is Prior-NN-IR, and C is the generated pathway.}
  \label{tab:compositional}
  \begin{tabular}{llrrrr}
    \toprule
    Modality & Variant    & Op-Set F1 (\%) & Build (\%) & STEP Export (\%) & plan cos \\
    \midrule
    STEP     & A          & 79.2           & 100.0      & 100.0            & 0.634    \\
             & B          & 76.6           & 100.0      & 99.97            & 0.534    \\
             & \textbf{C} & 64.4           & 46.7       & 46.0             & 0.499    \\
    Text     & A          & 53.7           & 98.1       & 98.1             & 0.342    \\
             & B          & 54.4           & 98.3       & 98.3             & 0.328    \\
             & \textbf{C} & 45.5           & 45.0       & 41.8             & 0.125    \\
    Image    & A          & 51.3           & 100.0      & 100.0            & 0.253    \\
             & B          & 40.0           & 99.97      & 99.97            & 0.267    \\
             & \textbf{C} & 41.2           & 62.4       & 60.0             & 0.226    \\
    Point    & A          & 43.8           & 97.6       & 97.6             & 0.160    \\
             & B          & 47.2           & 97.9       & 97.9             & 0.226    \\
             & \textbf{C} & 38.3           & 55.5       & 54.5             & 0.059    \\
    \bottomrule
  \end{tabular}
\end{table}

Retrieval remains at $97.6$--$100.0$~\% Build, whereas the generated pathway ranges from $45.0$ to $62.4$~\%. Holding out template families therefore does not remove the construction patterns exploited by nearest-neighbour retrieval.

\subsection{Construction-skeleton redundancy}
\label{app:skeleton}

The nearest retrieved neighbour has a median operation-set F1 of exactly $1.0$ against the query's reference plan. For a typical held-out query, a training sample from another template family therefore contains the same set of construction operations.

This result explains why family-level partitioning is insufficient as a proxy for construction novelty in this corpus. Template identity and construction structure are not equivalent: two parts can belong to different families while sharing the same operation skeleton. A benchmark intended to test compositional generalisation would need to define separation directly in construction space rather than only through generator-defined family labels.

The retrieval condition should therefore be interpreted as a strong reference rather than as a strict upper bound. Matching the operation set does not determine operation order, parameter values, semantic roles, or final geometry. The measured redundancy is consistent with broader observations that duplicate or near-duplicate training data can inflate evaluation scores~\cite{lee2022dedup}, and that nearest-neighbour retrieval can itself be a strong predictor~\cite{khandelwal2020knnlm}.

\subsection{Coverage and conditional fidelity}
\label{app:accountings}

Build rate measures whether a program reaches a valid solid, but it does not determine how well that solid matches the target. We therefore report two complementary F@1 accountings for STEP queries on the family-held-out split, using $F_{\mathrm{self}}=0.244$ as the empirical reference.

\begin{table}[htbp]
  \centering
  \footnotesize
  \caption{Two accountings of the same STEP comparison on the family-held-out
    split. \emph{Conditional} uses the 1,330 parts scoreable in both arms and
    reports a median. \emph{Unconditional} uses all 2,923 parts, assigns zero
    to non-scoreable outputs for this accounting only, and reports a mean.
    Percentages of $F_{\mathrm{self}}$ are shown in parentheses.}
  \label{tab:accountings}

  \setlength{\tabcolsep}{4pt}

  \begin{tabular}{
      >{\raggedright\arraybackslash}p{0.15\linewidth}
      >{\centering\arraybackslash}p{0.08\linewidth}
      >{\centering\arraybackslash}p{0.18\linewidth}
      >{\centering\arraybackslash}p{0.18\linewidth}
      >{\centering\arraybackslash}p{0.13\linewidth}
      >{\centering\arraybackslash}p{0.13\linewidth}
    }
    \toprule
    Accounting
     & $n$
     & Retrieved
     & Generated
     & Wins (Gen.:Ret.)
     & $p$                     \\
    \midrule

    \textbf{Conditional}
     & 1{,}330
     & 0.111 (45.6\%)
     & \textbf{0.162 (66.2\%)}
     & 982:347
     & $1.7\times10^{-70}$     \\

    \textbf{Unconditional}
     & 2{,}923
     & \textbf{0.099 (40.6\%)}
     & 0.066 (26.9\%)
     & 983:1{,}939
     & $4.1\times10^{-71}$     \\

    \bottomrule
  \end{tabular}
\end{table}

The two accountings answer different questions. On the jointly scoreable subset, the generated pathway has higher median F@1 and wins 982 of 1,329 non-tied paired comparisons. When all inputs are included, retrieval leads because it produces scoreable geometry much more often. The resulting reversal is therefore a coverage--fidelity trade-off rather than a contradiction between metrics.

This pattern is specific to the STEP condition considered here. For point-cloud queries, the conditional F@1 values are close ($0.022$ for generation and $0.020$ for retrieval), while retrieval retains a clear advantage under all-input accounting ($24.0$~\% versus $7.0$~\% of $F_{\mathrm{self}}$).

\subsection{Survivorship analysis}
\label{app:survivorship}

Conditioning on joint scoreability selects a subset of the evaluation data.
For the survivorship analysis, we consider the $2{,}922$ parts for which
retrieval provides an F@1 score. Among them, $1{,}330$ are also scoreable by
the generated pathway, while $1{,}592$ are not. The remaining one of the
$2{,}923$ test parts is scoreable by generation but not by retrieval, so it
cannot be included in this comparison.

Retrieval has a median F@1 of $0.1113$ on the $1{,}330$ jointly scoreable
parts and $0.1045$ on the $1{,}592$ generated-nonscoreable parts, a difference
of $0.0068$. The corresponding conditional gap between generation and
retrieval is $0.0503$. The observed ease shift is therefore about $14$~\% of
that gap.

The jointly scoreable subset is therefore somewhat easier for retrieval as
well. However, this shift is much smaller than the full conditional gap and
captures only one aspect of selection. We therefore report conditional
fidelity together with all-input coverage.

\section{External CAD-Recode Evaluation Details}
\label{app:external}

Section~\ref{sec:external_positioning} compares the MIRAGE point-cloud pathways with the released CAD-Recode checkpoint on externally authored Fusion 360 parts. This appendix documents the information conditions, the common evaluator, and additional checks on scoreability and measurement resolution.

\subsection{Information conditions and common evaluator}
\label{app:external_evaluator}

The comparison is intended as system-level external positioning rather than a
reproduction of CAD-Recode's published benchmark. Both systems are therefore
evaluated under the common protocol described below, and the reported numbers
should be interpreted only within this setting.

\begin{table}[htbp]
  \centering
  \footnotesize
  \caption{Information available to each pathway in the external point-cloud
    comparison.}
  \label{tab:external_information}

  \begin{tabular}{
      >{\raggedright\arraybackslash}p{0.23\linewidth}
      >{\raggedright\arraybackslash}p{0.53\linewidth}
      >{\centering\arraybackslash}p{0.12\linewidth}
    }
    \toprule
    Pathway
     & Conditioning at inference
     & Autonomous                                                   \\
    \midrule

    CAD-Recode
     & Point cloud $\rightarrow$ code
     & Yes                                                          \\

    MIRAGE generated plan
     & Point cloud $\rightarrow$ construction latent
    $\rightarrow$ plan $\rightarrow$ code
     & Yes                                                          \\

    MIRAGE Prior-NN-IR
     & Plan retrieved from the \textbf{training index} at inference
     & No                                                           \\

    \bottomrule
  \end{tabular}
\end{table}

The primary like-for-like comparison is between CAD-Recode and the autonomous
MIRAGE generated-plan pathway. Prior-NN-IR is retained as a retrieval-assisted
reference because it has access to training-set construction information that
CAD-Recode does not use.

All three outputs are passed through the same downstream evaluator:

\begin{enumerate}
  \item The same reference STEP file is used for every arm. Identity of the reference files and of the normalised meshes used for scoring was verified by hashing.
  \item STEP files are tessellated with a fixed dimensionless operator: linear deflection $\alpha L$ with $\alpha=10^{-6}$ and $L$ the largest bounding-box extent, and angular deflection $0.3$.
  \item Prediction and reference are independently normalised to the unit cube before geometric scoring, removing absolute position and scale from this comparison.
  \item Chamfer distance follows the released CAD-Recode demo: $8{,}192$ surface samples per shape, bidirectional mean squared nearest-neighbour distances, summed and multiplied by $1000$.
  \item Volumetric IoU is computed by mesh Boolean intersection and union.
  \item CD and IoU use their own scoreable subsets. Evaluator failures are reported separately rather than assigned a geometric score.
\end{enumerate}

This protocol differs from the internal geometry evaluation, which uses $1{,}024$ surface samples, reports Chamfer in mm$^2$, and uses F@1 at $1$~\% of the target bounding-box diagonal. The two Chamfer quantities are therefore reported separately and are not converted into one another.

The input point representations also differ. CAD-Recode follows its released preprocessing, which samples $8{,}192$ surface points and reduces them to $256$ points by deterministic farthest-point sampling. MIRAGE uses its deployed $1{,}024$-point normalised cloud. Equalising the point budget would require changing at least one released system and is therefore outside the scope of this comparison.

Because both prediction and reference are normalised to the unit cube, the comparison is insensitive to absolute scale. The known scale blindness of the MIRAGE point-cloud preprocessing therefore does not directly account for the geometric gap measured here.

\subsection{Tessellation and convergence}
\label{app:tessellation}

The chosen tessellation is compared with a ten-times-finer reference tessellation to estimate residual discretisation error. The median absolute IoU deviation is $2.5\times10^{-5}$ and the 95th percentile is $3.7\times10^{-4}$.

These deviations are substantially smaller than the IoU differences observed between systems, indicating that the reported ordering is not driven by the selected tessellation tolerance.

\subsection{Sampling-resolution analysis}
\label{app:sampling_floor}

Finite surface sampling imposes a part-dependent resolution floor on Chamfer distance. Across the $400$ parts, this floor varies by roughly a factor of fifty. We therefore compare each paired system difference with the larger sampling floor of the corresponding part pair.

No paired difference falls below this resolution threshold: $0$ of $222$ pairs in the generated-plan comparison and $0$ of $385$ pairs in the Prior-NN-IR comparison satisfy $|\Delta\mathrm{CD}|$ below the larger sampling floor.

A second normalisation expresses CD relative to each part's own floor, $r=\mathrm{CD}/\mathrm{floor}$. CAD-Recode has a median ratio of approximately $1.24$--$1.29$, while the two MIRAGE pathways are $208$ and $782$ times above their corresponding floors. The paired differences are $+199$ $[+172,+226]$ and $+766$ $[+556,+1061]$.

Together with the consistent IoU ordering, these checks indicate that the external fidelity gap is not explained by the finite sampling resolution of the Chamfer metric.

\subsection{Metric scoreability and failure breakdown}
\label{app:external_scoreability}

External coverage is defined as the number of inputs for which the common evaluator can obtain measurable geometry. This criterion is stricter than STEP export. The generated-plan pathway exports $232$ STEP files, of which $226$ are measurable; Prior-NN-IR exports $395$, of which $392$ are measurable.

\begin{table}[htbp]
  \centering
  \scriptsize
  \setlength{\tabcolsep}{3pt}

  \caption{External measurable-geometry coverage over all 400 Fusion 360
    inputs. The count columns report parts measurable by both systems, by
    CAD-Recode only, by MIRAGE only, and by neither. McNemar tests are exact.
    CADR denotes CAD-Recode.}
  \label{tab:ext_coverage}

  \begin{tabular}{lccrrrrl}
    \toprule
    Pathway
     & CADR
     & MIRAGE
     & Both
     & CADR only
     & MIR. only
     & Neither
     & $p$                 \\
    \midrule

    Generated plan
     & 392/400
     & \textbf{226/400}
     & 222
     & 170
     & 4
     & 4
     & $3.2\times10^{-45}$ \\

    Prior-NN-IR
     & 392/400
     & \textbf{392/400}
     & 385
     & 7
     & 7
     & 1
     & \textbf{1.0}        \\

    \bottomrule
  \end{tabular}
\end{table}

For the autonomous generated-plan pathway, $174$ of $400$ inputs do not reach measurable geometry: $168$ do not export STEP, five export STEP files that tessellate to no triangles, and one exports a solid containing no faces. The remaining $226$ measurable outputs still show substantially lower fidelity than CAD-Recode, so the external deficit is not explained by coverage alone.

Across the two MIRAGE pathways, nine outputs pass all five internal validity gates but remain unscoreable in the external evaluator: eight re-tessellate to zero triangles and one contains no faces. These cases illustrate the distinction between executable validity and metric scoreability.

Boolean failures affect IoU but not CD. They occur for $14$ of $226$ generated-plan outputs, $3$ of $392$ Prior-NN-IR outputs, and $13$ of $392$ CAD-Recode outputs.

\subsection{Additional paired statistics}
\label{app:external_paired}

\begin{table}[htbp]
  \centering
  \scriptsize
  \setlength{\tabcolsep}{3pt}

  \caption{External paired fidelity on jointly scoreable subsets.
    $\Delta\mathrm{CD}
      =\mathrm{CD}_{\text{MIRAGE}}-\mathrm{CD}_{\text{CAD-Recode}}$,
    so positive values favour CAD-Recode.
    $\Delta\mathrm{IoU}
      =\mathrm{IoU}_{\text{MIRAGE}}-\mathrm{IoU}_{\text{CAD-Recode}}$,
    so positive values favour MIRAGE. The reported $\Delta$ is the paired
    median of per-part differences. Exact ties are excluded from the sign
    tests. CADR denotes CAD-Recode.}
  \label{tab:ext_paired}

  \begin{tabular}{lrrrrlrl}
    \toprule
    Pairing
     & $n$
     & MIRAGE
     & CADR
     & $\Delta$
     & 95\% CI
     & MIR. wins
     & $p$                  \\
    \midrule

    Gen.-plan CD
     & 222
     & 55.78
     & 0.117
     & $+55.64$
     & $[+49.29,+63.84]$
     & 1/222
     & $6.6\times10^{-65}$  \\

    Gen.-plan IoU
     & 200
     & 0.0604
     & 0.9226
     & $-0.785$
     & $[-0.810,-0.751]$
     & 5/200
     & $1.2\times10^{-50}$  \\

    Prior-NN-IR CD
     & 385
     & 17.59
     & 0.187
     & $+17.28$
     & $[+15.51,+19.84]$
     & \textbf{5/385}
     & $1.8\times10^{-105}$ \\

    Prior-NN-IR IoU
     & 369
     & 0.1621
     & 0.9426
     & $-0.684$
     & $[-0.704,-0.658]$
     & 10/369
     & $2.8\times10^{-91}$  \\

    \bottomrule
  \end{tabular}
\end{table}

Prior-NN-IR and CAD-Recode have the same marginal measurable-geometry
coverage, $392$ of $400$ parts each. On paired coverage, $385$ parts are
measurable for both systems, with seven discordant successes in each direction
and exact McNemar $p=1.0$. Of these $385$ parts, $369$ are jointly
IoU-scoreable. On that paired subset, the median IoU is $0.1621$ for
Prior-NN-IR and $0.9426$ for CAD-Recode.

These paired medians differ from the marginal medians computed on each
system's own scoreable subset. The corresponding marginal IoU medians are
$0.157$ for Prior-NN-IR and $0.938$ for CAD-Recode because the two scoreable
subsets are not identical.

The common-success subset does not explain the poor MIRAGE result.
CAD-Recode's median CD is $0.117$ on the $222$ parts jointly scoreable with
the generated-plan arm, compared with $0.186$ over all $392$ parts for which
CAD-Recode has a CD score. The subset reconstructed by MIRAGE is therefore,
if anything, easier for CAD-Recode as well.

Overall, the external comparison places the current MIRAGE point-cloud
pathway well below the specialised CAD-Recode system in geometric fidelity.
This result concerns system-level point-cloud reconstruction and does not test
the internal representation question addressed by the construction-conditioning
experiments in Sections~\ref{sec:rq1} and~\ref{sec:ablation_prefix}.

\section{Implementation and Reproducibility Details}
\label{app:implementation}

\subsection{Corpus composition and splits}
\label{app:corpus}

Samples are divided into four complexity tiers according to operation count and
Boolean complexity: L1 ($\approx 3$~\%, one to two operations), L2
($\approx 17$~\%, three to five operations), L3 ($\approx 55$~\%, multiple
sketches with Boolean subtraction or intersection), and L4 ($\approx 25$~\%,
revolve, array, shell, loft, or multi-body Boolean operations). This
stratification is preserved across the data splits.

The main experiments use $25{,}000$ training, $2{,}500$ validation, and
$2{,}500$ test samples per modality. Retrieval indices are constructed from
training samples only; validation and test programs are excluded.

Reference programs are executed and kernel-validated when the corpus is built.
Consequently, evaluation failures reported in this paper arise from generated
programs rather than from invalid reference programs.

\subsection{Hardware and quantisation}
\label{app:hardware}

All experiments were run on a single NVIDIA RTX~5060~Ti with $16$~GB of memory.
The two language-model stages use Qwen2.5-Coder-1.5B~\cite{yang2025qwen25} with
4-bit NF4 weights, double quantisation, and bfloat16 computation following
QLoRA~\cite{dettmers2023qlora}. LoRA adapters~\cite{hu2021lora} are trained on
top of the quantised base model. Stages~1 and~2 train the smaller encoders and
MLPs in bfloat16 without 4-bit quantisation.

At inference, the plan and code adapters are loaded sequentially. The plan
adapter is unloaded after plan generation and the CUDA cache is cleared before
the code adapter is loaded. This arrangement was used to keep inference within
the memory available on a single GPU.

\subsection{Per-stage training settings}
\label{app:stage_settings}

\begin{center}
  \footnotesize
  \setlength{\tabcolsep}{4pt}

  \begin{tabular}{
      >{\raggedright\arraybackslash}p{0.24\linewidth}
      >{\raggedright\arraybackslash}p{0.66\linewidth}
    }
    \toprule
    Stage & Settings                                                        \\
    \midrule

    1 Representation alignment
          & Batch 8, gradient accumulation 4 (effective batch 32), AdamW at
    $10^{-4}$, cosine decay, 5 epochs, $\tau=0.07$, maximum text/IR
    length 128/256, 1,024 points, and rare operations oversampled
    $2\times$.                                                              \\

    2 Latent prior fitting
          & Batch 48, AdamW at $10^{-4}$, cosine decay, 8 epochs,
    $\tau=0.07$, with all three loss weights set to 1.0.                    \\

    3 Plan-decoder adaptation
          & Batch 1, gradient accumulation 16, AdamW at $2\times10^{-4}$,
    cosine decay, 3 epochs, maximum length 1,536, LoRA $r=8$,
    $\alpha=16$, and dropout 0.05.                                          \\

    3b Cross-modal continuation
          & Continuation of Stage~3 for 1 epoch at $10^{-5}$ on
    2,000 training rows.                                                    \\

    4 Code-decoder training
          & Batch 1, gradient accumulation 8, AdamW at $2\times10^{-4}$,
    1 epoch, maximum length 1,536, LoRA $r=16$, $\alpha=32$,
    and dropout 0.05.                                                       \\

    4b Predicted-plan continuation
          & Continuation of Stage~4 for 3 epochs at $5\times10^{-5}$ using
    the 70/30 reference/predicted-plan mixture.                             \\

    \bottomrule
  \end{tabular}
\end{center}

The maximum sequence length is fixed at 1,536 tokens for the controlled decoder
comparisons. Memory pressure is handled through gradient accumulation rather
than by shortening the context, which would change the information available
to the decoder.

Two details of the training provenance are useful for reproducing the reported
runs. First, the predicted-plan portion of Stage~4b was generated from the
Stage~3 plan-decoder checkpoint, rather than the later cross-modal continuation,
using 1,000 training rows filtered to grammar-valid plans. Second, the
family-held-out experiment retrains the full pipeline from scratch on its own
partition while keeping the same hyperparameter settings.

\subsection{Decoding, checkpoint selection, and repair}
\label{app:decoding}

Unless stated otherwise, generation is greedy and single-shot ($N=1$). No
geometry-based candidate selection is applied in the main experiments. Runs
with $N>1$ are reported separately and labelled accordingly. Paired conditions
use the same decoding settings.

Checkpoint selection differs between the direct-latent and plan-mediated
controlled runs. The direct-latent trainer restores its best validation
checkpoint, whereas the plan-mediated run is evaluated at its final step. The
resulting 3,000-versus-3,125 update difference and its implications are
reported in Appendix~\ref{app:budget}.

The deterministic repair rules described in Appendix~\ref{app:failures} are
applied once before execution. They are fixed string-level rewrites; execution
errors are not fed back into the decoder and no program is regenerated on the
basis of kernel feedback.

\subsection{Single-run scope}
\label{app:single_run}

Each trained condition reported in the paper corresponds to one training run.
Seeds were fixed and recorded, but the experiments do not provide an estimate
of seed-to-seed variance. Statistical comparisons between trained conditions
therefore quantify variation over evaluated samples conditional on the trained
checkpoints. This is particularly relevant to the small coverage difference
between B1a and B2-Pred discussed in Section~\ref{sec:direct_latent}.

A multi-seed replication of that comparison was prepared but could not be
completed on the available hardware. The manuscript therefore makes no claim
about the stability of the observed coverage difference across training seeds.

\subsection{Implementation-specific limitations}
\label{app:defects}

Table~\ref{tab:defects} collects implementation details that affect the scope
of particular results. They are separated from the methodological limitations
in Section~\ref{sec:discussion_limitations} because each can, in principle,
be changed without altering the overall construction-mediated formulation.

\begin{table}[htbp]
  \centering
  \footnotesize
  \caption{Implementation-specific limitations and the changes required to
    address them.}
  \label{tab:defects}

  \begin{tabular}{
      >{\raggedright\arraybackslash}p{0.28\linewidth}
      >{\raggedright\arraybackslash}p{0.32\linewidth}
      >{\raggedright\arraybackslash}p{0.30\linewidth}
    }
    \toprule
    Implementation detail
     & Consequence
     & Possible correction                                                     \\
    \midrule

    The point-cloud input is divided by its maximum radius before encoding
     & Absolute scale is removed before the encoder; the produced-to-target
    bounding-box ratio is 0.695 internally and 0.339 on the cross-source
    set.
     & Provide the pre-normalisation scale as an additional input and retrain
    from the alignment stage.                                                  \\

    The point-cloud path contains a train--inference prompt inconsistency in
    the cross-modal continuation
     & Point-cloud plan quality is confounded by this prompt mismatch; its
    magnitude and direction are not isolated.
     & Correct the prompt and retrain the cross-modal continuation.            \\

    \texttt{Prog-Op-F1} matches operation identifiers wherever they occur
     & 61.6\% of matched identifiers are parameter names inside string
    literals; the metric's absolute value should not be interpreted as
    operation-call accuracy.
     & Restrict scoring to canonical operation invocations or operation
    families; this requires rescoring but not retraining.                      \\

    Each trained condition is represented by a single run
     & Small differences between trained configurations include unmeasured
    seed-to-seed variation.
     & Repeat the comparisons supporting the headline claims across multiple
    training seeds.                                                            \\

    Stages~3, 4, and 4b use STEP rows and their adapters are reused across
    modalities
     & The multimodal results combine modality effects with a decoder-training
    distribution that is not symmetric across modalities.
     & Train or continue the code decoder with modality-balanced conditioning,
    or evaluate modality-specific decoder adaptations.                         \\

    The STEP face, edge, and relation streams contain no independent per-entity
    records in the present corpus
     & Reported STEP results effectively depend on the 50-dimensional global
    descriptor (Appendix~\ref{app:step_effective}); they do not establish
    the performance of a learned entity-level B-Rep encoder.
     & Export per-entity descriptors during corpus construction and retrain
    from the alignment stage.                                                  \\

    \bottomrule
  \end{tabular}
\end{table}

\clearpage

\bibliographystyle{elsarticle-num}
\bibliography{references}

\end{document}